\documentclass{article}

\PassOptionsToPackage{numbers,sort&compress}{natbib}
\usepackage[eandd, final]{neurips_2026}

\usepackage[utf8]{inputenc} 
\usepackage[T1]{fontenc}    
\usepackage{hyperref}       
\usepackage{url}            
\usepackage{booktabs}       
\usepackage{amsfonts}       
\usepackage{nicefrac}       
\usepackage{microtype}      
\usepackage{xcolor}         

\usepackage{amsmath}
\usepackage{amssymb}
\usepackage{mathtools}
\usepackage{amsthm}

\usepackage{threeparttable}  
\usepackage{multirow}        

\usepackage{amsfonts}       
\usepackage{lipsum}
\usepackage{graphicx}

\usepackage{algorithm}
\usepackage{algorithmic}

\usepackage{amsfonts}
\usepackage[table]{xcolor}
\usepackage{multicol}
\usepackage{tabularx}
\usepackage{subfigure}
\usepackage{subcaption}
\usepackage{caption}
\usepackage{wrapfig}

\newcommand{\pmci}[1]{\textsubscript{\scriptsize$\pm$#1}}

\title{\textsc{SynBench}: A Benchmark for Differentially Private Text Generation}

\author{
  \textbf{Yidan~Sun}$^{1}$\thanks{Corresponding author: \texttt{y.sun1@imperial.ac.uk}},
  \textbf{Viktor~Schlegel}$^{1}$,
  \textbf{Srinivasan~Nandakumar}$^{1}$,
  \textbf{Iqra~Zahid}$^{2}$,
  \textbf{Yuping~Wu}$^{3}$, \\
  \textbf{Yulong~Wu}$^{3}$,
  \textbf{Hao~Li}$^{4}$,
  \textbf{Jie~Zhang}$^{5}$,
  \textbf{Warren~Del-Pinto}$^{3}$,
  \textbf{Goran~Nenadic}$^{3}$,\\
  \textbf{Siew~Kei~Lam}$^{6}$,
  \textbf{Anil~Anthony~Bharath}$^{1,4}$ \\[4pt]
  $^{1}$Imperial College London, Imperial Global Singapore \\
  $^{2}$Microsoft, London, UK \quad
  $^{3}$University of Manchester \quad
  $^{4}$Imperial College London \\
  $^{5}$A*STAR \quad
  $^{6}$Nanyang Technological University
}

\begin{document}

\maketitle

\begin{abstract}
  Synthetic text generation with Differential Privacy (DP) guarantees emerges as a principled approach that can enable the sharing of sensitive datasets across institutional and regulatory boundaries, while bounding the risks of re-identification and membership inference. LLM-based methods deliver promising results; however, comparisons are exacerbated by differing evaluation setups and ``private'' datasets, potential pre-training contamination is not considered and guarantees are not verified with DP audits.
  To advance this field, we introduce a unified evaluation framework with standardised utility and fidelity metrics and privacy audits, encompassing nine curated datasets that capture domain-specific complexities such as technical jargon, long-context dependencies, and specialised document structures. In a large-scale empirical study, we benchmark LLM-based state-of-the-art DP text generators of varying sizes (between 1--8B). 
  Our results indicate that DP synthetic text generation remains an unsolved challenge, with quality deteriorating more as the private datasets deviate further from the generators' pre-training corpora. 
  Our analyses suggest that synthetic data quality may be overestimated when LLMs have been pre-trained---without DP---on portions of the ``private'' data to be generated, and our membership inference (MIA) auditing setup for synthetic text reveals measurable membership signals even under tight privacy budgets.
  Finally, we note that under the ``public pre-training, private generation'' paradigm, the nominal $(\varepsilon,\delta)$ guarantee does not account for exposure during pre-training, motivating contamination-aware evaluation of DP synthetic text generation. 
  Code and data splits: \url{https://github.com/ImperialGlobalSingapore/synbench}
\end{abstract}

\section{Introduction}

The proliferation of modern language technology in the form of Large (neural) Language Models (LLMs) is governed by scaling laws which posit that model performance improves substantially with increased training data and size \citep{kaplan2020scaling}. As open-~\citep{dubey2024llama} and closed-source models~\citep{OpenAI2023GPT-4Report} continue to scale to hundreds of billions of parameters, they require ever-increasing amounts of diverse, high-quality training data to fuel the growth. However, publicly available data represents a finite resource that is rapidly being exhausted \cite{muennighoff2023scaling}.

While vast troves of valuable data exist within specialised domains---including healthcare records~\cite{Johnson2023MIMIC-IVDataset}, legal documents~\cite{chalkidis2019large} or financial transactions and proprietary business communications~\cite{wu2023bloomberggpt}---these datasets remain inaccessible to the broader research community due to privacy, confidentiality, and regulatory constraints. Making such domain-specific data available could alleviate the data bottleneck and help adopt language technology to aid experts in those domains~\cite{DellAcqua2023NavigatingQuality}.

Direct training on sensitive datasets, however, poses fundamental privacy risks. Recent work has demonstrated that language models can memorise and subsequently leak verbatim training examples~\citep{Carlini2021ExtractingModels}, potentially exposing sensitive information. 
Synthetic data generation~\cite{Schick2021GeneratingModels,Li2023Team:PULSARModels} has emerged as a promising alternative, offering the possibility of capturing the statistical properties of private datasets without directly exposing individual records. Yet, even synthetic data generation is not immune to privacy concerns, as recent work has shown that sensitive information can be recovered from synthetic datasets generated by language models trained on private data~\citep{meeus2025canary}.

\begin{figure}[t]
  \centering
  \begin{minipage}{0.5\textwidth}
    \centering
    \includegraphics[width=\linewidth]
    {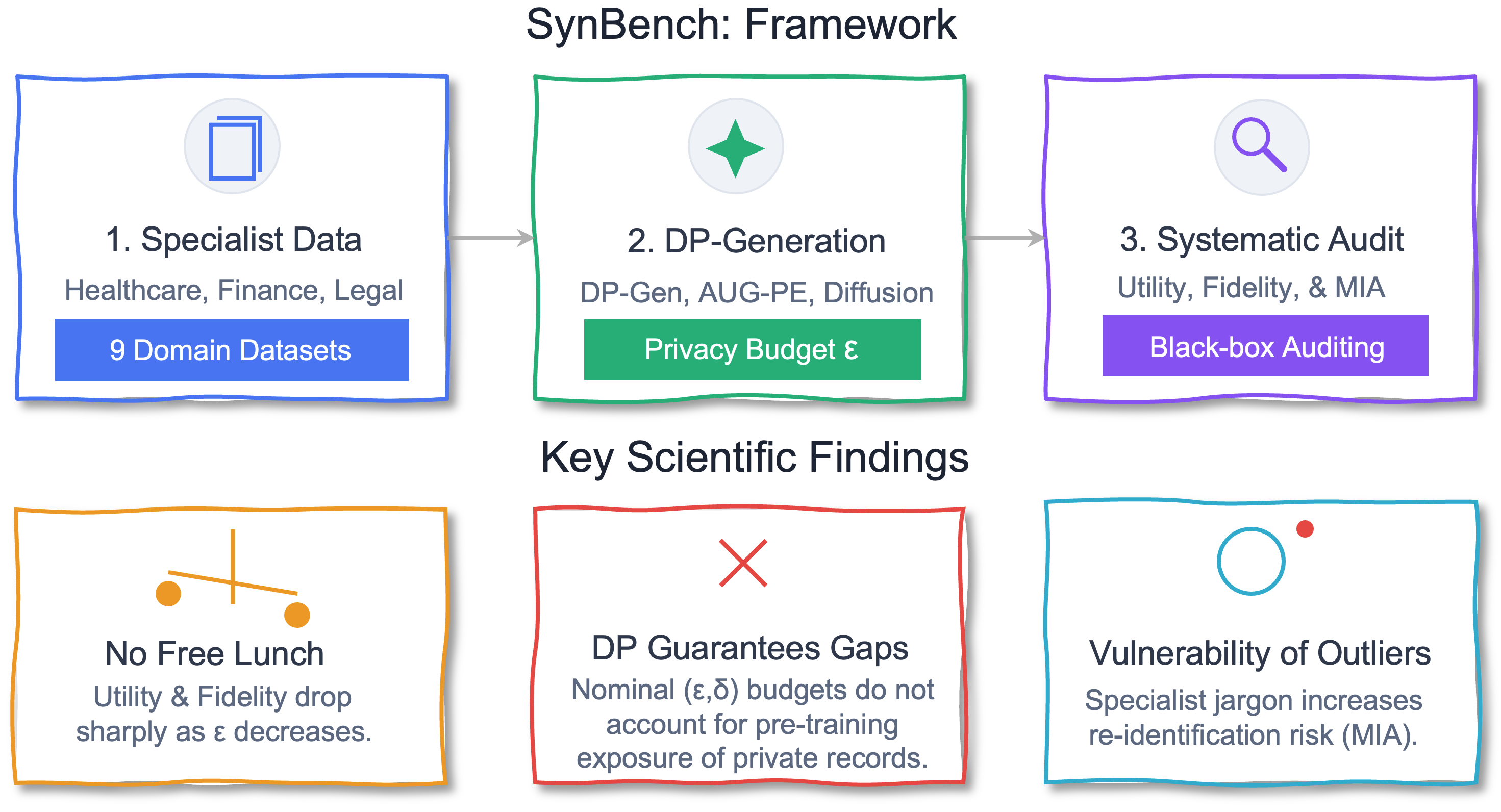}
    \caption{SynBench Overview. The framework integrates (1) Specialist Data, (2) DP-Generation, and (3) Systematic Audit. 
    }
    \label{img:frameworkillustration}
  \end{minipage}
  \hfill
  \begin{minipage}{0.49\textwidth}
    \centering
    \includegraphics[width=\linewidth]{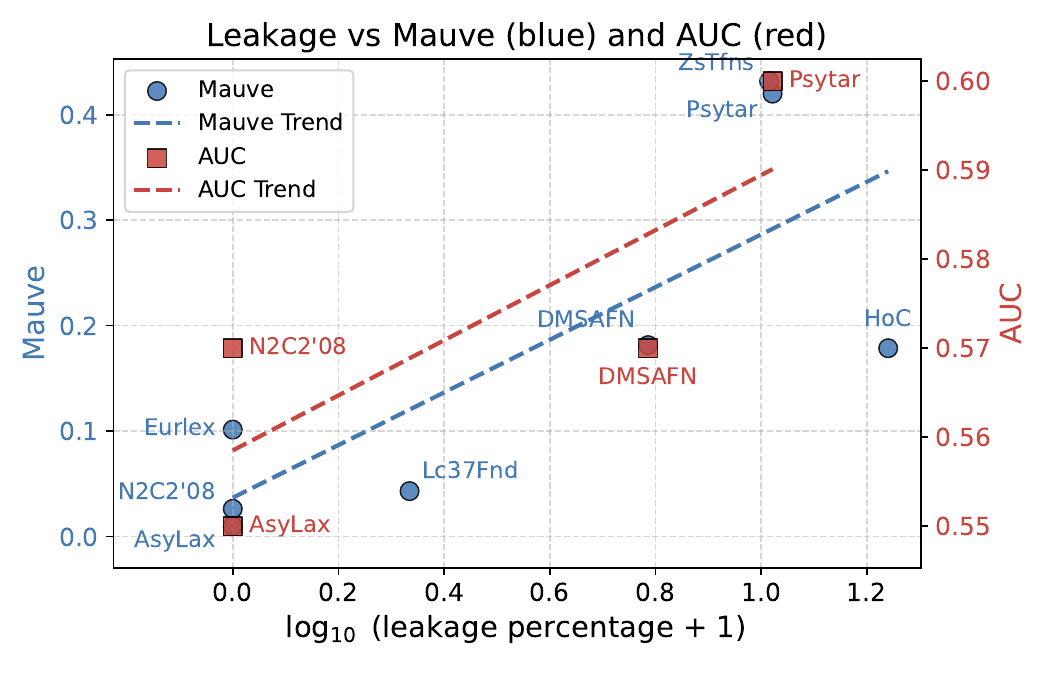}
    \caption{\textbf{No free lunch.} Across datasets, higher estimated pre-training overlap goes with higher synthetic quality and MIA AUC (MAUVE; Spearman $\rho = 0.76$, $p = 0.04$).
    }
    \label{img:leakage_corr}
  \end{minipage}
\end{figure}

To effectively address such concerns, differential privacy~(DP), the gold standard for privacy-preserving data analysis~\citep{dwork2014algorithmic}, can be used to provide formal mathematical guarantees that bound the risk of private data leakage. While DP has been successfully applied to synthetic text generation~\cite{Yue2023SyntheticRecipe,Xie2024DifferentiallyText,Mattern2022DifferentiallySharing}, the field remains fragmented and faces several critical challenges that limit its practical adoption.

First, existing work relies on a pick-and-choose approach to datasets and evaluation metrics, with \emph{no standardised methodology for assessment}. This exacerbates meaningful comparisons across methods, identification of systematic limitations, and progress tracking of the field \citep{schlegel2025generating}. 
Second, current research predominantly focuses on simple, open-domain datasets such as product~\citep{Mattern2022DifferentiallySharing} or paper reviews~\citep{Xie2024DifferentiallyText}, \emph{failing to address the domain-specific challenges} present in high-stakes applications. 
These challenges include specialised terminology, complex document structures, long-form content, and domain-specific quality requirements~\citep{huang2025survey,zhang2025siren}.

The use of pre-trained models for data synthesis introduces additional complications: many benchmark datasets used to evaluate differentially private synthetic data generation may have been encountered during the pre-training of LLM-based generation methods \citep{schlegel2025generating,sun2025evaluating}. This raises concerns of potentially overestimating generation quality at specified privacy budgets, as models may be leveraging memorised information rather than truly learning from the private datasets~\citep{Tramer2024Position:Pretraining}. In simple terms, \emph{access to private data during pre-training is not accounted for} in the privacy guarantees. Compounding this issue is the \emph{lack of rigorous empirical verification for privacy leakage}: For example, \citet{Yue2023SyntheticRecipe} only provide anecdotal leakage evidence of canary instances, while \citet{Xie2024DifferentiallyText} only evaluate data privacy leakage of downstream models trained on synthetic data. 
Consequently, the extent of privacy leakage due to pre-training
contamination, which the nominal $(\varepsilon,\delta)$ guarantee does not cover,
remains largely unknown.

In this work we introduce \textbf{SynBench}, a \textbf{benchmark and auditing framework} for differentially private (DP) synthetic text generation that evaluates methods for their \textbf{utility, fidelity, and privacy leakage} using a standardised protocol. 
SynBench is comprised of \textbf{nine datasets}, spanning high-stakes and domain-specific settings with specialized terminology, long-form structure, and non-trivial label spaces. This enables us to compare methods in controlled settings, which is different otherwise, given the fragmented evaluation landscape of current state of the art. 
Using SynBench, we run a \textbf{large-scale empirical study} of four state-of-the-art DP generation pipelines and pair it with a \textbf{membership-inference-based DP audit} tailored to synthetic text. 
Across datasets and methods, we find a consistent and practically relevant message: \textbf{current approaches do not offer a reliable privacy--utility win} and our MIA audits reveal \textbf{measurable membership signal even at tight privacy budgets}.
Data ``contamination'', i.e., when ``private'' evaluation data appear in the training corpora of the underlying LLM generator, can explain both these observations. First, this can inflate measured utility and fidelity: For OLMo-2 we observe a small but consistent rise in downstream utility when training a classifier on synthetic data generated from increasingly ``leaked'' private samples---up to 3.4\% utility gain. Second, the nominal $(\varepsilon,\delta)$ guarantee bounds only the influence of a record's inclusion in the private data; it does not account for content the generator was already exposed to during pre-training, which might explain the observed exceedance of $\epsilon$ in our experiments, compared to DP guarantees.

\section{Background \& Related Work}

\textbf{Preliminaries} Differential Privacy (DP) provides a formal mathematical framework for quantifying and limiting the privacy risk incurred when releasing data derived from sensitive sources. A mechanism \(\mathcal{M}\) satisfies \((\epsilon, \delta)\)-differential privacy if, for any two neighbouring datasets \(x\) and \(x'\) differing by a single record, and for any subset \(S\) of outputs, the inequality
\[
\Pr[\mathcal{M}(x) \in S] \leq e^{\epsilon} \Pr[\mathcal{M}(x') \in S] + \delta
\]
holds~\cite{Dwork20026}. This formalism bounds the advantage of an adversary attempting a Membership Inference Attack (MIA), making DP a key foundation for privacy-preserving data generation. Importantly, any operations on the outputs of the DP mechanism preserve this guarantee (post-processing property).

\textbf{DP Synthetic Text Generation} Existing methods fall broadly into two
paradigms. \emph{DP training} approaches~\cite{Yue2023SyntheticRecipe,
Mattern2022DifferentiallySharing,meeus2025canary} fine-tune generative
models conditioned on ``control codes'' (typically class labels) with
DP-SGD~\cite{Abadi2016DeepPrivacy}, which clips per-example gradients and
adds noise; by post-processing, prompting the trained model yields DP
synthetic data. The same recipe applies to text diffusion
models~\cite{ochs2025private} and to ACTG-ARL~\cite{hu2026actg}, which
first synthesises tabular features with DP and then generates text
conditioned on them. \emph{DP inference} methods~\cite{koga2024privacy,
Flemings2024DifferentiallyModels,Tang2023Privacy-PreservingGeneration}
instead privatise generation, e.g., via PATE~\cite{Papernot2018ScalablePATE};
since private data are accessed at every step, the privacy loss grows per
token, which is impractical for large-scale synthesis
(Appendix~\ref{app:dp-inference}). Exceptions are AUG-PE~\cite{Xie2024DifferentiallyText},
which iteratively evolves random samples towards the private distribution
under a privatised fitness function, and EPSVec~\cite{banayeeanzade2026epsvec},
which privatises a single activation-space \emph{dataset vector} once and
injects it during decoding, so its cost does not grow with the number of
samples. We evaluate representatives of each design point: DP training
(DP-Gen, DP-Diffusion), private evolution (AUG-PE), and privatised
steering (EPSVec). We exclude ``rewriting'' approaches that map private to
synthetic records one-to-one, as they either break the privacy guarantee
or require local DP~\cite{Habernal2021WhenDetail}, an unnecessarily strong
assumption for centralised releases by trusted curators.

\textbf{Research Gaps in Evaluation} Despite the steady developments in DP synthetic text generation, fundamental evaluation challenges remain unanswered. First, existing work predominantly evaluates on open-domain datasets (sentiment analysis~\cite{Yue2023SyntheticRecipe,Mattern2022DifferentiallySharing} or paper abstracts~\cite{Xie2024DifferentiallyText}) that are likely included in foundation model pre-training corpora. 
Such contamination may lead to privacy leakage that is not accounted for by the DP guarantee, as models may have accessed and memorized benchmark ``private'' data
during pre-training, and may inflate synthetic data quality estimates.
This is a critical concern~\cite{Tramer2024Position:Pretraining}, that remains unaddressed in current literature. Second, empirical privacy evaluation is often incomplete: many works rely solely on formal guarantees without empirical validation~\cite{Mattern2022DifferentiallySharing,Flemings2024DifferentiallyGeneration}, while others substitute privacy evaluation of the synthesized data itself for simpler analyses~\cite{Yue2023SyntheticRecipe,Xie2024DifferentiallyText}. Most critically, the field lacks evaluation on realistic, domain-specific datasets from high-stakes applications where synthetic data generation would provide the greatest value. 
\section{Benchmark Design}

\textbf{Evaluation Baselines}
We consider four representative baselines that instantiate distinct design points for differentially private (DP) text synthesis. 
\textbf{AUG-PE} \cite{Xie2024DifferentiallyText} is an API-based differentially private synthetic text generator that estimates privatized feature statistics from the sensitive corpus and iteratively queries a large language model (LLM) via its API to generate text matching those statistics. 
\textbf{DP-transformers} (which is referred as DP-Gen) \cite{dp-transformers} represents end-to-end neural generation under DP by training Transformer language or seq2seq models with DP-SGD, i.e., per-example gradient clipping and additive noise during optimization, followed by standard decoding to sample synthetic texts. 
\textbf{DP-Diffusion} (which is referred as \textbf{DP-Dfs}) \cite{ochs2025private} is a diffusion-based DP text generator that learns a reverse denoising process for text and enforces privacy during training (e.g., via DP-SGD), enabling iterative denoising-based sampling while typically exhibiting sensitivity to strong privacy noise in small-data regimes.
\textbf{EPSVec}~\cite{banayeeanzade2026epsvec} steers a pretrained LLM
by injecting a privatized \emph{dataset vector}, an activation-space
direction encoding the shift from public priors to the private corpus,
into its hidden states; since the vector is privatized only once, the
privacy cost does not grow with the number of generated samples.

\textbf{Domain-specific datasets}
For data, we use four healthcare datasets: \textsc{HoC} \cite{DBLP:journals/bioinformatics/BakerSGAHSK16} (cancer hallmark identification in scientific literature), \textsc{N2C2'08}\cite{uzuner2009recognizing} (obesity and co-morbidity recognition in clinical discharge summaries), \textsc{PsyTAR} \cite{Zolnoori2019} (adverse drug effect detection in social media posts), and \textsc{Mimic} \cite{mimic2023} (triage of discharge summaries, with ICD codes grouped into top-level categories).
Additionally, we evaluate three financial datasets: \textsc{DMSAFN}\footnote{ \url{https://huggingface.co/datasets/Daniel-ML/sentiment-analysis-for-financial-news-v2}} and \textsc{Lc37Fnd} \cite{drinkall2025financialregression} (both financial news), and \textsc{ZsTfns}\footnote{\url{https://huggingface.co/datasets/zeroshot/twitter-financial-news-sentiment}} (financial tweets), all used for sentiment classification.
We also include two legal datasets: \textsc{AsyLex} \cite{barale-etal-2023-automated}, which introduces a legal reasoning task, and \textsc{EurLex} \cite{chalkidis2019large}, a multi-label classification dataset of legal documents covering diverse legal concepts. These datasets broaden the scope of our benchmark beyond healthcare and finance.
The challenges associated with these datasets are summarized in Table~\ref{tab:datasets}. Our benchmark features long documents (\textsc{N2C2'08}, \textsc{Lc37Fnd}, \textsc{AsyLex}); gated-access datasets (\textsc{PsyTAR}, \textsc{N2C2'08}, and \textsc{Mimic}); small datasets that complicate model training (\textsc{N2C2'08}); and multi-label datasets with many classes (\textsc{HoC}, \textsc{N2C2'08}, \textsc{PsyTAR}, \textsc{MIMIC} and \textsc{EurLex}).

\begin{wraptable}{r}{0.5\textwidth}
  \centering
  \caption{Benchmark characteristics: Number of labels ($|C|$), dataset size ($|D|$), avg/p90 token length $|\overline L|$ and data access. HF: huggingface. DUA: data usage agreement.}
  \label{tab:datasets}
  \resizebox{\linewidth}{!}{\begin{tabular}{lcccc}
\toprule
\textbf{Dataset} & \textbf{$|C|$} & \textbf{$|D|$} & \textbf{$|\overline{L}|$} & \textbf{Access mechanism} \\
\midrule
\textsc{HoC} & 10 & 10301 & 37/58 & free on HF\\
\textsc{N2C2'08} & 16 & 620 & 1985/3070 & DUA \& manual approval \\
\textsc{PsyTAR} & 7 & 5102 & 19/34 & DUA \\
\textsc{Mimic} & 21 & 4971 & 2726/3951 & DUA \& credentialed access\\
\textsc{DMSAFN} & 3 & 3876 & 30/50 & free on HF \\
\textsc{Lc37Fnd} & 3 & 5620 & 836/1510 & free on HF \\
\textsc{ZsTfns} & 3 & 9543 & 25/41 & free on HF \\
\textsc{AsyLax} & 3 & 7999 & 2326/3442& free on HF\\
\textsc{Eurlex} & 8 & 4942 & 266/563& free on HF\\
\bottomrule
\end{tabular}

}
\end{wraptable}

\textbf{Evaluation Protocol} Our benchmark framework provides a comprehensive evaluation of \textit{utility}, \textit{fidelity}, and \textit{privacy leakage} across varying privacy budgets (i.e., different $\epsilon$-levels). 

\textbf{Utility} assesses the practical usefulness of synthetic data for downstream application tasks. Specifically, we train multiple classification models on the synthetic data and evaluate their micro-F1 on a held-out portion of the original dataset.
Each classifier is trained with three random seeds, and we report the mean and 95\% confidence interval over seeds, which reflects variability in classifier training for a fixed synthetic dataset. 
This setup enables us to determine how well synthetic data can support real-world analytical objectives.

\textbf{Fidelity} measures the degree of similarity between synthetic and original data across multiple linguistic dimensions. We use MAUVE~\cite{Pillutla2021MAUVE:Frontiers} to quantify semantic alignment, while the divergence in named entity recognition (NER) distributions in sentence level and text length statistics between the original and synthetic corpora serves to evaluate lexical and structural similarity. 
These metrics offer a complementary perspective, particularly in practical scenarios where the real data is unavailable or restricted. 

\textbf{Privacy leakage} is assessed using Membership Inference Attacks (MIAs) under a realistic black-box scenario. 
In Machine Learning, MIA is a privacy attack in which an adversary aims to determine whether a given data point was part of a machine learning model's training dataset. 
Most of existing MIA methods assumed that attackers have access to the trained model or its output logits~\cite{carlini2019secret,shi2023detecting}. 
However, for synthetic data, we ask: \emph{How effectively does synthetic text generated by LLMs protect the privacy of the private dataset?}
Therefore, we consider a realistic setup where the adversary only has access to the synthetic dataset and no knowledge of the underlying text generation model or training data~\cite{meeus2025canary}. 

Crucially, we study privacy leakage on \emph{real} samples without inserting canaries. We consider a realistic but strong adversary who has access to authentic candidate records. 
Following standard practice, we assume a strong adversary with full knowledge of the target record and of the data distribution, from which it can draw auxiliary data to train reference models. To match the synthetic-data release setting, the adversary observes only the released
synthetic dataset and has no access to the generator model.
We further focus on \emph{outlier} target records, which are plausibly more vulnerable to membership inference. Concretely, we select outliers from each private dataset as MIA targets (see below). This setup quantifies how much LLM-generated synthetic text can expose private information, and we evaluate leakage empirically under varying privacy budgets $\epsilon$ against the corresponding theoretical MIA upper bounds.

\textbf{Membership Inference Attack}
\label{MIA}
Our proposed pipeline consists of three main stages:

\noindent\emph{Stage 1: Outlier Detection.}  
Given a private dataset $D$, we embed each sample using the same encoder as in AUG‑PE \cite{Xie2024DifferentiallyText}. From this embedding space, we identify extreme outliers---including the top 1\% farthest points and Local Outlier Factor (LoF) outliers \cite{breunig2000lof}---as our attack targets, denoted by $D_{\text{out}}$. While such outliers seldom dominate real-world attack targets, this selection simulates a worst-case scenario for auditing.

\noindent\emph{Stage 2: Dataset Partitioning and Reference Construction.}  
After removing $D_{\text{out}}$ from $D$, the remaining data are split into a private subset $D_{\text{prv}}$ and an auxiliary subset $D_{\text{aux}}$ in a 1:2 ratio, maintaining identical data distributions.  
We then sample 100 subsets $\{D_{\text{prv}}^i\}_{i=1}^{100}$ from $D_{\text{prv}}$, each with a 50\% chance of including a randomly selected outlier from $D_{\text{out}}$. The corresponding synthetic datasets $\{\hat{D}_{\text{prv}}^i\}_{i=1}^{100}$  will be generated using $\{D_{\text{prv}}^i\}_{i=1}^{100}$ by various generators under different $\epsilon$ values, respectively.
Concurrently, we generate $M=4$ reference auxiliary sets $\{D_{\text{ref}}^1,\dots,D_{\text{ref}}^4\}$, each a duplicate of $D_{\text{aux}}$. For each target $x\in D_{\text{out}}$, exactly two of these references include $x$, while the other two exclude it, ensuring balanced membership signals. We then train separate n‑gram language models on each $D_{\text{ref}}^i$~\cite{jm3}.

\noindent\emph{Stage 3: Attack Signal Computation.}  
For each generated synthetic set $\hat{D}_{\text{prv}}^i$ and target record $x_i$, we train an n‑gram model on $\hat{D}_{\text{prv}}^i$  and use it to estimate $P_{\mathrm{ng}}(x_i)$.
We also compute the average probability $\bar{P}_{\mathrm{ref}}(x_i)$ across the four reference models.  
The final MIA attack score is thus defined as:
$\Delta P(x_i) = P_{\mathrm{ng}}(x_i) - \bar{P}_{\mathrm{ref}}(x_i)$.
A higher $\Delta P(x_i)$ suggests a greater likelihood that $x_i$ was included in $D_{\text{prv}}^i$.

This pipeline efficiently simulates a realistic black-box MIA, requiring only one synthetic dataset per attack trial while providing statistically grounded membership inference.  
We then use the computed attack scores together with the true membership labels to generate Receiver Operating Characteristic (ROC) curves. 
From these curves, we derive a key metric of attack performance: Area Under the Curve (AUC), which reflects the overall discriminative ability of the attack by measuring the area under the ROC. 

\noindent\textbf{Experiment Setup} While there is no consensus on how much
$\varepsilon$ is ``enough''~\cite{lee2011much}, high values provide
increasingly weaker protection~\cite{dwork2019differential}. In practice
(e.g., as recommended by NIST~\cite{NISTDPChallenges}),
$\varepsilon \leq 5$ or even $\varepsilon \leq 1$ is considered a strong
privacy guarantee, which guides our choice of
$\varepsilon \in \{0.5, 1, 2, 4\}$. For \textbf{AUG-PE}, \textbf{DP-Gen}, and \textbf{EPSVec}, we use
Llama models (Llama-3.2-1B, Llama-3.2-3B, and Llama-3.1-8B) as
the LLM backbone across all datasets. 
For \textbf{DP-dfs}, we use GENIE~\cite{lin2023text}, the best-performing backbone reported in~\cite{ochs2025private}. Further implementation details are provided in
Appendix~\ref{app:additional-implementation}.
\section{Empirical Results \& Analysis}
\label{sec:experiments}

\begin{table*}[t]
\centering
\setlength{\tabcolsep}{3pt}
\resizebox{\textwidth}{!}{%
\begin{tabular}{|l|ccccc|ccccc|ccccc|}
\hline
\multicolumn{16}{|c|}{\textsc{Dataset HoC} (\textbf{Random}: 3.7, \textbf{Majority}: 9.1)} \\ \hline
\cellcolor{gray!20} & \multicolumn{5}{c|}{\cellcolor{gray!20}\textbf{BERT-large (original=71.7{\scriptsize$\pm$2.2})}} & \multicolumn{5}{c|}{\cellcolor{gray!20}\textbf{BioClinicalBERT (original=67.6{\scriptsize$\pm$5.8})}} & \multicolumn{5}{c|}{\cellcolor{gray!20}\textbf{BiomedBERT (original=74.0{\scriptsize$\pm$0.7})}} \\
Method & $\epsilon=\infty$ & $\epsilon=0.5$ & $\epsilon=1$ & $\epsilon=2$ & $\epsilon=4$ & $\epsilon=\infty$ & $\epsilon=0.5$ & $\epsilon=1$ & $\epsilon=2$ & $\epsilon=4$ & $\epsilon=\infty$ & $\epsilon=0.5$ & $\epsilon=1$ & $\epsilon=2$ & $\epsilon=4$ \\
DP-Gen & \textbf{48.2} & \textcolor{gray}{11.2} & \textcolor{gray}{9.7} & 13.0 & 11.6 & \textbf{45.3} & \textcolor{gray}{10.0} & \textcolor{gray}{9.5} & 11.9 & 12.2 & \textbf{53.0} & 12.7 & 12.6 & 17.0 & 16.8 \\[-3pt]
 & {\scriptsize $\pm$3.9} & {\scriptsize $\pm$4.7} & {\scriptsize $\pm$3.2} & {\scriptsize $\pm$4.0} & {\scriptsize $\pm$0.5} & {\scriptsize $\pm$3.6} & {\scriptsize $\pm$2.6} & {\scriptsize $\pm$3.7} & {\scriptsize $\pm$2.7} & {\scriptsize $\pm$0.5} & {\scriptsize $\pm$1.7} & {\scriptsize $\pm$3.6} & {\scriptsize $\pm$4.3} & {\scriptsize $\pm$8.9} & {\scriptsize $\pm$2.0} \\
AUG-PE & \textcolor{gray}{11.0} & \textcolor{gray}{3.8} & \textcolor{gray}{5.4} & \textcolor{gray}{4.4} & \textcolor{gray}{7.1} & \textcolor{gray}{10.4} & \textcolor{gray}{2.9} & \textcolor{gray}{6.7} & \textcolor{gray}{5.2} & \textcolor{gray}{8.3} & 16.9 & \textcolor{gray}{7.0} & \textcolor{gray}{6.9} & \textcolor{gray}{5.5} & \textcolor{gray}{8.2} \\[-3pt]
 & {\scriptsize $\pm$6.5} & {\scriptsize $\pm$0.4} & {\scriptsize $\pm$4.9} & {\scriptsize $\pm$1.6} & {\scriptsize $\pm$2.5} & {\scriptsize $\pm$6.2} & {\scriptsize $\pm$3.3} & {\scriptsize $\pm$3.4} & {\scriptsize $\pm$2.3} & {\scriptsize $\pm$1.4} & {\scriptsize $\pm$7.0} & {\scriptsize $\pm$4.5} & {\scriptsize $\pm$2.0} & {\scriptsize $\pm$1.9} & {\scriptsize $\pm$6.7} \\
DP-Dfs & 40.1 & 18.3 & 18.2 & 18.1 & 16.9 & 40.0 & 18.8 & 19.0 & 18.5 & 18.4 & 44.1 & \textbf{27.6} & 23.7 & \textbf{30.1} & \textbf{30.6} \\[-3pt]
 & {\scriptsize $\pm$3.9} & {\scriptsize $\pm$8.2} & {\scriptsize $\pm$3.9} & {\scriptsize $\pm$3.8} & {\scriptsize $\pm$0.7} & {\scriptsize $\pm$3.9} & {\scriptsize $\pm$7.6} & {\scriptsize $\pm$4.7} & {\scriptsize $\pm$4.5} & {\scriptsize $\pm$2.7} & {\scriptsize $\pm$3.1} & {\scriptsize $\pm$23.0} & {\scriptsize $\pm$26.2} & {\scriptsize $\pm$7.1} & {\scriptsize $\pm$2.2} \\
EPSVec & 29.1 & \textbf{20.7} & \textbf{22.2} & \textbf{22.9} & \textbf{22.9} & 29.2 & \textbf{20.9} & \textbf{23.7} & \textbf{24.2} & \textbf{23.8} & 34.9 & 25.7 & \textbf{26.9} & 27.0 & 28.5 \\[-3pt]
 & {\scriptsize $\pm$2.7} & {\scriptsize $\pm$4.5} & {\scriptsize $\pm$3.5} & {\scriptsize $\pm$2.5} & {\scriptsize $\pm$4.0} & {\scriptsize $\pm$5.4} & {\scriptsize $\pm$3.9} & {\scriptsize $\pm$1.2} & {\scriptsize $\pm$3.7} & {\scriptsize $\pm$3.6} & {\scriptsize $\pm$3.6} & {\scriptsize $\pm$1.0} & {\scriptsize $\pm$7.9} & {\scriptsize $\pm$7.2} & {\scriptsize $\pm$6.2} \\
\hline
\multicolumn{16}{|c|}{\textsc{Dataset N2C2'08} (\textbf{Random}: 46.1, \textbf{Majority}: 53.2)} \\ \hline
\cellcolor{gray!20} & \multicolumn{5}{c|}{\cellcolor{gray!20}\textbf{LongFormer-large (original=82.6{\scriptsize$\pm$14.5})}} & \multicolumn{5}{c|}{\cellcolor{gray!20}\textbf{Clinical-BigBird (original=87.4{\scriptsize$\pm$2.4})}} & \multicolumn{5}{c|}{\cellcolor{gray!20}\textbf{Clinical-LongFormer (original=77.9{\scriptsize$\pm$43.5})}} \\
Method & $\epsilon=\infty$ & $\epsilon=0.5$ & $\epsilon=1$ & $\epsilon=2$ & $\epsilon=4$ & $\epsilon=\infty$ & $\epsilon=0.5$ & $\epsilon=1$ & $\epsilon=2$ & $\epsilon=4$ & $\epsilon=\infty$ & $\epsilon=0.5$ & $\epsilon=1$ & $\epsilon=2$ & $\epsilon=4$ \\
DP-Gen & \textbf{68.9} & \textcolor{gray}{54.0} & \textcolor{gray}{52.0} & \textcolor{gray}{52.5} & \textcolor{gray}{50.2} & \textbf{75.5} & 58.2 & \textcolor{gray}{52.9} & \textcolor{gray}{53.7} & 55.2 & \textbf{75.6} & \textcolor{gray}{46.6} & \textcolor{gray}{47.9} & 57.4 & \textcolor{gray}{47.8} \\[-3pt]
 & {\scriptsize $\pm$20.4} & {\scriptsize $\pm$15.2} & {\scriptsize $\pm$12.7} & {\scriptsize $\pm$5.4} & {\scriptsize $\pm$16.1} & {\scriptsize $\pm$1.5} & {\scriptsize $\pm$5.2} & {\scriptsize $\pm$4.8} & {\scriptsize $\pm$6.5} & {\scriptsize $\pm$2.6} & {\scriptsize $\pm$6.3} & {\scriptsize $\pm$11.4} & {\scriptsize $\pm$24.9} & {\scriptsize $\pm$3.8} & {\scriptsize $\pm$15.2} \\
AUG-PE & \textcolor{gray}{53.7} & \textcolor{gray}{52.1} & 61.2 & 56.1 & 58.0 & 58.6 & \textcolor{gray}{51.3} & 55.6 & 55.5 & \textcolor{gray}{54.6} & 59.9 & \textcolor{gray}{50.2} & 57.3 & 57.0 & 56.6 \\[-3pt]
 & {\scriptsize $\pm$9.3} & {\scriptsize $\pm$4.8} & {\scriptsize $\pm$0.3} & {\scriptsize $\pm$5.7} & {\scriptsize $\pm$5.3} & {\scriptsize $\pm$3.0} & {\scriptsize $\pm$7.8} & {\scriptsize $\pm$2.6} & {\scriptsize $\pm$5.1} & {\scriptsize $\pm$8.1} & {\scriptsize $\pm$4.1} & {\scriptsize $\pm$23.3} & {\scriptsize $\pm$3.2} & {\scriptsize $\pm$4.5} & {\scriptsize $\pm$10.9} \\
DP-Dfs & 61.5 & 56.0 & \textbf{64.3} & \textbf{61.0} & \textbf{63.3} & 62.1 & \textcolor{gray}{38.0} & \textbf{63.5} & \textbf{65.8} & \textbf{65.6} & 56.9 & \textcolor{gray}{41.6} & \textbf{61.3} & \textbf{65.0} & \textbf{67.2} \\[-3pt]
 & {\scriptsize $\pm$3.1} & {\scriptsize $\pm$10.3} & {\scriptsize $\pm$3.6} & {\scriptsize $\pm$1.2} & {\scriptsize $\pm$4.1} & {\scriptsize $\pm$3.4} & {\scriptsize $\pm$14.0} & {\scriptsize $\pm$13.8} & {\scriptsize $\pm$8.7} & {\scriptsize $\pm$6.5} & {\scriptsize $\pm$8.8} & {\scriptsize $\pm$13.7} & {\scriptsize $\pm$9.8} & {\scriptsize $\pm$1.8} & {\scriptsize $\pm$16.8} \\
EPSVec & \textcolor{gray}{24.9} & \textbf{56.7} & \textcolor{gray}{34.2} & \textcolor{gray}{15.9} & \textcolor{gray}{13.9} & \textcolor{gray}{40.3} & \textbf{59.0} & \textcolor{gray}{42.4} & \textcolor{gray}{24.0} & \textcolor{gray}{10.7} & \textcolor{gray}{24.8} & \textcolor{gray}{\textbf{53.6}} & \textcolor{gray}{30.5} & \textcolor{gray}{30.9} & \textcolor{gray}{16.6} \\[-3pt]
 & {\scriptsize $\pm$8.9} & {\scriptsize $\pm$2.4} & {\scriptsize $\pm$1.5} & {\scriptsize $\pm$12.1} & {\scriptsize $\pm$18.0} & {\scriptsize $\pm$30.1} & {\scriptsize $\pm$3.8} & {\scriptsize $\pm$14.8} & {\scriptsize $\pm$25.2} & {\scriptsize $\pm$23.8} & {\scriptsize $\pm$32.3} & {\scriptsize $\pm$16.1} & {\scriptsize $\pm$41.9} & {\scriptsize $\pm$24.9} & {\scriptsize $\pm$17.0} \\
\hline
\multicolumn{16}{|c|}{\textsc{Dataset PsyTAR} (\textbf{Random}: 25.4, \textbf{Majority}: 41.8)} \\ \hline
\cellcolor{gray!20} & \multicolumn{5}{c|}{\cellcolor{gray!20}\textbf{BERT-base (original=78.6{\scriptsize$\pm$0.6})}} & \multicolumn{5}{c|}{\cellcolor{gray!20}\textbf{BERT-large (original=79.5{\scriptsize$\pm$1.6})}} & \multicolumn{5}{c|}{\cellcolor{gray!20}\textbf{DeBERTa-xlarge (original=82.1{\scriptsize$\pm$1.1})}} \\
Method & $\epsilon=\infty$ & $\epsilon=0.5$ & $\epsilon=1$ & $\epsilon=2$ & $\epsilon=4$ & $\epsilon=\infty$ & $\epsilon=0.5$ & $\epsilon=1$ & $\epsilon=2$ & $\epsilon=4$ & $\epsilon=\infty$ & $\epsilon=0.5$ & $\epsilon=1$ & $\epsilon=2$ & $\epsilon=4$ \\
DP-Gen & \textbf{66.8} & \textcolor{gray}{36.4} & \textcolor{gray}{36.8} & \textcolor{gray}{39.9} & \textcolor{gray}{40.5} & \textbf{67.5} & \textcolor{gray}{36.3} & \textcolor{gray}{36.2} & \textcolor{gray}{41.2} & \textcolor{gray}{40.4} & \textbf{71.3} & \textcolor{gray}{37.9} & \textcolor{gray}{41.5} & \textcolor{gray}{42.0} & \textcolor{gray}{43.1} \\[-3pt]
 & {\scriptsize $\pm$2.3} & {\scriptsize $\pm$4.3} & {\scriptsize $\pm$4.4} & {\scriptsize $\pm$1.0} & {\scriptsize $\pm$3.3} & {\scriptsize $\pm$2.8} & {\scriptsize $\pm$0.5} & {\scriptsize $\pm$2.5} & {\scriptsize $\pm$4.1} & {\scriptsize $\pm$1.5} & {\scriptsize $\pm$1.2} & {\scriptsize $\pm$1.8} & {\scriptsize $\pm$2.9} & {\scriptsize $\pm$2.8} & {\scriptsize $\pm$3.7} \\
AUG-PE & 63.1 & \textcolor{gray}{\textbf{42.2}} & \textbf{52.0} & \textbf{54.8} & \textbf{56.8} & 65.4 & \textcolor{gray}{\textbf{42.5}} & \textbf{55.4} & \textbf{59.4} & \textbf{58.0} & 67.8 & \textcolor{gray}{\textbf{43.4}} & \textbf{55.0} & \textbf{60.9} & \textbf{61.3} \\[-3pt]
 & {\scriptsize $\pm$0.5} & {\scriptsize $\pm$2.6} & {\scriptsize $\pm$3.3} & {\scriptsize $\pm$2.2} & {\scriptsize $\pm$2.8} & {\scriptsize $\pm$2.1} & {\scriptsize $\pm$0.8} & {\scriptsize $\pm$3.2} & {\scriptsize $\pm$1.6} & {\scriptsize $\pm$3.2} & {\scriptsize $\pm$2.4} & {\scriptsize $\pm$7.9} & {\scriptsize $\pm$3.4} & {\scriptsize $\pm$6.6} & {\scriptsize $\pm$4.3} \\
DP-Dfs & 63.6 & -- & \textcolor{gray}{18.2} & \textcolor{gray}{24.5} & \textcolor{gray}{25.3} & 64.2 & -- & \textcolor{gray}{16.3} & \textcolor{gray}{19.7} & \textcolor{gray}{21.9} & 65.9 & -- & \textcolor{gray}{26.3} & \textcolor{gray}{25.7} & \textcolor{gray}{22.7} \\[-3pt]
 & {\scriptsize $\pm$1.3} &  & {\scriptsize $\pm$21.1} & {\scriptsize $\pm$11.6} & {\scriptsize $\pm$4.2} & {\scriptsize $\pm$3.6} &  & {\scriptsize $\pm$12.4} & {\scriptsize $\pm$7.0} & {\scriptsize $\pm$5.1} & {\scriptsize $\pm$2.8} &  & {\scriptsize $\pm$6.5} & {\scriptsize $\pm$4.5} & {\scriptsize $\pm$4.1} \\
EPSVec & \textcolor{gray}{42.7} & \textcolor{gray}{35.6} & \textcolor{gray}{42.9} & \textcolor{gray}{40.1} & 44.5 & \textcolor{gray}{44.0} & \textcolor{gray}{39.3} & 45.6 & \textcolor{gray}{43.6} & 47.8 & 50.1 & \textcolor{gray}{42.0} & 45.9 & \textcolor{gray}{44.3} & 51.5 \\[-3pt]
 & {\scriptsize $\pm$1.1} & {\scriptsize $\pm$3.2} & {\scriptsize $\pm$3.8} & {\scriptsize $\pm$1.2} & {\scriptsize $\pm$5.2} & {\scriptsize $\pm$7.1} & {\scriptsize $\pm$4.0} & {\scriptsize $\pm$5.2} & {\scriptsize $\pm$4.0} & {\scriptsize $\pm$5.0} & {\scriptsize $\pm$1.3} & {\scriptsize $\pm$5.7} & {\scriptsize $\pm$6.2} & {\scriptsize $\pm$3.8} & {\scriptsize $\pm$2.3} \\
\hline
\multicolumn{16}{|c|}{\textsc{Dataset Mimic} (\textbf{Random}: 48.8, \textbf{Majority}: 52.5)} \\ \hline
\cellcolor{gray!20} & \multicolumn{5}{c|}{\cellcolor{gray!20}\textbf{BiomedBERT (original=76.3{\scriptsize$\pm$1.2})}} & \multicolumn{5}{c|}{\cellcolor{gray!20}\textbf{BioClinicalBERT (original=74.8{\scriptsize$\pm$0.2})}} & \multicolumn{5}{c|}{\cellcolor{gray!20}\textbf{BiomedRobertaBase (original=73.4{\scriptsize$\pm$0.9})}} \\
Method & $\epsilon=\infty$ & $\epsilon=0.5$ & $\epsilon=1$ & $\epsilon=2$ & $\epsilon=4$ & $\epsilon=\infty$ & $\epsilon=0.5$ & $\epsilon=1$ & $\epsilon=2$ & $\epsilon=4$ & $\epsilon=\infty$ & $\epsilon=0.5$ & $\epsilon=1$ & $\epsilon=2$ & $\epsilon=4$ \\
DP-Gen & \textbf{70.7} & \textcolor{gray}{52.2} & \textbf{62.3} & 59.6 & 63.5 & \textbf{70.1} & \textcolor{gray}{50.2} & \textbf{64.3} & 63.5 & 62.8 & \textbf{69.4} & \textcolor{gray}{45.4} & \textbf{62.3} & 63.2 & 61.7 \\[-3pt]
 & {\scriptsize $\pm$1.8} & {\scriptsize $\pm$10.1} & {\scriptsize $\pm$3.8} & {\scriptsize $\pm$5.1} & {\scriptsize $\pm$3.5} & {\scriptsize $\pm$2.8} & {\scriptsize $\pm$7.3} & {\scriptsize $\pm$0.6} & {\scriptsize $\pm$1.9} & {\scriptsize $\pm$1.1} & {\scriptsize $\pm$1.5} & {\scriptsize $\pm$17.0} & {\scriptsize $\pm$1.1} & {\scriptsize $\pm$0.8} & {\scriptsize $\pm$7.6} \\
AUG-PE & 64.7 & 62.1 & 56.2 & 61.9 & \textbf{64.9} & 60.6 & 63.6 & 57.2 & \textbf{63.7} & \textbf{66.0} & 59.3 & 60.0 & 55.8 & 59.7 & \textbf{66.3} \\[-3pt]
 & {\scriptsize $\pm$0.2} & {\scriptsize $\pm$0.6} & {\scriptsize $\pm$4.2} & {\scriptsize $\pm$6.7} & {\scriptsize $\pm$0.3} & {\scriptsize $\pm$4.6} & {\scriptsize $\pm$1.9} & {\scriptsize $\pm$0.6} & {\scriptsize $\pm$1.1} & {\scriptsize $\pm$2.2} & {\scriptsize $\pm$10.9} & {\scriptsize $\pm$4.0} & {\scriptsize $\pm$7.2} & {\scriptsize $\pm$6.9} & {\scriptsize $\pm$1.2} \\
DP-Dfs & 66.8 & 54.4 & \textcolor{gray}{53.1} & \textcolor{gray}{44.7} & \textcolor{gray}{43.5} & 64.1 & \textcolor{gray}{52.0} & \textcolor{gray}{49.4} & \textcolor{gray}{44.3} & \textcolor{gray}{46.0} & \textcolor{gray}{39.3} & \textcolor{gray}{51.9} & 54.3 & \textcolor{gray}{52.1} & 55.0 \\[-3pt]
 & {\scriptsize $\pm$3.5} & {\scriptsize $\pm$11.7} & {\scriptsize $\pm$3.1} & {\scriptsize $\pm$6.2} & {\scriptsize $\pm$12.1} & {\scriptsize $\pm$6.5} & {\scriptsize $\pm$13.2} & {\scriptsize $\pm$8.3} & {\scriptsize $\pm$8.1} & {\scriptsize $\pm$1.8} & {\scriptsize $\pm$14.2} & {\scriptsize $\pm$13.9} & {\scriptsize $\pm$10.0} & {\scriptsize $\pm$10.7} & {\scriptsize $\pm$15.2} \\
EPSVec & 60.0 & \textbf{64.6} & \textcolor{gray}{50.0} & \textbf{63.5} & 59.4 & 62.7 & \textbf{64.7} & \textcolor{gray}{48.3} & 61.9 & \textcolor{gray}{51.8} & 60.0 & \textbf{61.5} & 55.4 & \textbf{64.3} & 62.2 \\[-3pt]
 & {\scriptsize $\pm$7.8} & {\scriptsize $\pm$2.8} & {\scriptsize $\pm$6.6} & {\scriptsize $\pm$1.0} & {\scriptsize $\pm$4.5} & {\scriptsize $\pm$6.1} & {\scriptsize $\pm$2.5} & {\scriptsize $\pm$16.9} & {\scriptsize $\pm$4.1} & {\scriptsize $\pm$9.2} & {\scriptsize $\pm$3.0} & {\scriptsize $\pm$5.8} & {\scriptsize $\pm$15.7} & {\scriptsize $\pm$5.2} & {\scriptsize $\pm$5.3} \\
\hline
\end{tabular}}
\caption{Random/majority guess baselines, and downstream classification micro-F1 scores for three models per dataset, trained on synthetic data from Llama-3.2-1B across different privacy budgets ($\epsilon$): biomedical datasets. Each value is the mean over 3 seeds (small type: $\pm$ the 95\% confidence interval (t-distribution)). ``--'': no result (DP-Dfs settings whose synthetic data were unusable). Bold: the best method in each column; grey: not significantly above Majority (paired bootstrap over test examples, one-sided, Benjamini-Hochberg-corrected $p \geq 0.05$).}
\label{tab:utility-llama-bio}
\end{table*}

\begin{table*}[t]
\centering
\setlength{\tabcolsep}{3pt}
\resizebox{\textwidth}{!}{%
\begin{tabular}{|l|ccccc|ccccc|ccccc|}
\hline
\multicolumn{16}{|c|}{\textsc{Dataset DMSAFN} (\textbf{Random}: 30.5, \textbf{Majority}: 41.1)} \\ \hline
\cellcolor{gray!20} & \multicolumn{5}{c|}{\cellcolor{gray!20}\textbf{ProsusAI-finbert (original=79.6{\scriptsize$\pm$6.4})}} & \multicolumn{5}{c|}{\cellcolor{gray!20}\textbf{finbert-tone (original=73.4{\scriptsize$\pm$2.8})}} & \multicolumn{5}{c|}{\cellcolor{gray!20}\textbf{FinancialBERT (original=92.0{\scriptsize$\pm$3.9})}} \\
Method & $\epsilon=\infty$ & $\epsilon=0.5$ & $\epsilon=1$ & $\epsilon=2$ & $\epsilon=4$ & $\epsilon=\infty$ & $\epsilon=0.5$ & $\epsilon=1$ & $\epsilon=2$ & $\epsilon=4$ & $\epsilon=\infty$ & $\epsilon=0.5$ & $\epsilon=1$ & $\epsilon=2$ & $\epsilon=4$ \\
DP-Gen & \textbf{69.4} & 43.2 & \textbf{49.8} & 44.3 & 48.1 & \textbf{67.2} & \textcolor{gray}{38.9} & \textbf{47.4} & \textcolor{gray}{37.6} & \textcolor{gray}{38.4} & \textbf{81.4} & 51.8 & \textbf{60.1} & 53.7 & 56.5 \\[-3pt]
 & {\scriptsize $\pm$22.9} & {\scriptsize $\pm$2.6} & {\scriptsize $\pm$5.8} & {\scriptsize $\pm$17.8} & {\scriptsize $\pm$12.0} & {\scriptsize $\pm$3.8} & {\scriptsize $\pm$2.7} & {\scriptsize $\pm$1.1} & {\scriptsize $\pm$5.1} & {\scriptsize $\pm$13.9} & {\scriptsize $\pm$3.9} & {\scriptsize $\pm$3.4} & {\scriptsize $\pm$8.7} & {\scriptsize $\pm$10.4} & {\scriptsize $\pm$3.7} \\
AUG-PE & 46.4 & 50.2 & 45.4 & 48.5 & 47.7 & 43.9 & 44.8 & 42.1 & 43.1 & 44.2 & 52.6 & 51.1 & 51.6 & 54.4 & 53.4 \\[-3pt]
 & {\scriptsize $\pm$1.8} & {\scriptsize $\pm$15.5} & {\scriptsize $\pm$3.7} & {\scriptsize $\pm$1.3} & {\scriptsize $\pm$1.9} & {\scriptsize $\pm$0.4} & {\scriptsize $\pm$4.6} & {\scriptsize $\pm$0.3} & {\scriptsize $\pm$0.7} & {\scriptsize $\pm$1.6} & {\scriptsize $\pm$1.7} & {\scriptsize $\pm$13.0} & {\scriptsize $\pm$8.9} & {\scriptsize $\pm$9.3} & {\scriptsize $\pm$8.7} \\
DP-Dfs & 62.6 & -- & -- & \textbf{65.3} & 60.2 & 60.6 & -- & -- & \textbf{55.3} & \textcolor{gray}{41.4} & 64.7 & -- & -- & \textbf{67.9} & 62.6 \\[-3pt]
 & {\scriptsize $\pm$1.2} &  &  & {\scriptsize $\pm$23.1} & {\scriptsize $\pm$33.3} & {\scriptsize $\pm$5.6} &  &  & {\scriptsize $\pm$18.8} & {\scriptsize $\pm$11.5} & {\scriptsize $\pm$6.8} &  &  & {\scriptsize $\pm$3.2} & {\scriptsize $\pm$11.7} \\
EPSVec & 51.4 & \textbf{54.0} & 49.4 & 51.7 & \textbf{70.4} & \textcolor{gray}{32.5} & \textbf{45.4} & \textcolor{gray}{40.6} & 44.2 & \textbf{54.7} & 66.0 & \textbf{64.8} & 54.2 & 60.2 & \textbf{71.2} \\[-3pt]
 & {\scriptsize $\pm$22.3} & {\scriptsize $\pm$4.3} & {\scriptsize $\pm$9.0} & {\scriptsize $\pm$4.4} & {\scriptsize $\pm$5.0} & {\scriptsize $\pm$15.8} & {\scriptsize $\pm$5.8} & {\scriptsize $\pm$4.5} & {\scriptsize $\pm$5.1} & {\scriptsize $\pm$5.0} & {\scriptsize $\pm$11.9} & {\scriptsize $\pm$16.9} & {\scriptsize $\pm$5.8} & {\scriptsize $\pm$13.5} & {\scriptsize $\pm$7.6} \\
\hline
\multicolumn{16}{|c|}{\textsc{Dataset Lc37Fnd} (\textbf{Random}: 41.3, \textbf{Majority}: 50.2)} \\ \hline
\cellcolor{gray!20} & \multicolumn{5}{c|}{\cellcolor{gray!20}\textbf{ProsusAI-finbert (original=86.7{\scriptsize$\pm$0.1})}} & \multicolumn{5}{c|}{\cellcolor{gray!20}\textbf{distilroberta (original=91.1{\scriptsize$\pm$1.8})}} & \multicolumn{5}{c|}{\cellcolor{gray!20}\textbf{FinancialBERT (original=82.7{\scriptsize$\pm$1.9})}} \\
Method & $\epsilon=\infty$ & $\epsilon=0.5$ & $\epsilon=1$ & $\epsilon=2$ & $\epsilon=4$ & $\epsilon=\infty$ & $\epsilon=0.5$ & $\epsilon=1$ & $\epsilon=2$ & $\epsilon=4$ & $\epsilon=\infty$ & $\epsilon=0.5$ & $\epsilon=1$ & $\epsilon=2$ & $\epsilon=4$ \\
DP-Gen & \textbf{77.1} & \textbf{56.2} & \textbf{58.4} & \textbf{58.1} & \textbf{59.0} & \textbf{80.4} & \textbf{55.8} & \textbf{63.8} & \textbf{64.3} & \textbf{66.4} & \textbf{75.7} & \textbf{54.7} & 55.5 & 57.3 & \textbf{58.8} \\[-3pt]
 & {\scriptsize $\pm$3.2} & {\scriptsize $\pm$4.9} & {\scriptsize $\pm$4.4} & {\scriptsize $\pm$14.2} & {\scriptsize $\pm$5.0} & {\scriptsize $\pm$1.9} & {\scriptsize $\pm$3.1} & {\scriptsize $\pm$2.5} & {\scriptsize $\pm$3.8} & {\scriptsize $\pm$7.4} & {\scriptsize $\pm$4.2} & {\scriptsize $\pm$1.2} & {\scriptsize $\pm$4.8} & {\scriptsize $\pm$2.1} & {\scriptsize $\pm$2.8} \\
AUG-PE & \textcolor{gray}{52.0} & \textcolor{gray}{43.9} & \textcolor{gray}{43.3} & \textcolor{gray}{42.2} & \textcolor{gray}{46.2} & 55.3 & \textcolor{gray}{44.8} & \textcolor{gray}{46.3} & \textcolor{gray}{49.0} & \textcolor{gray}{48.5} & \textcolor{gray}{53.9} & \textcolor{gray}{48.1} & \textcolor{gray}{46.7} & \textcolor{gray}{51.6} & \textcolor{gray}{48.6} \\[-3pt]
 & {\scriptsize $\pm$15.8} & {\scriptsize $\pm$11.6} & {\scriptsize $\pm$6.6} & {\scriptsize $\pm$10.6} & {\scriptsize $\pm$5.4} & {\scriptsize $\pm$22.7} & {\scriptsize $\pm$4.5} & {\scriptsize $\pm$5.3} & {\scriptsize $\pm$1.8} & {\scriptsize $\pm$7.7} & {\scriptsize $\pm$14.0} & {\scriptsize $\pm$3.7} & {\scriptsize $\pm$7.0} & {\scriptsize $\pm$16.8} & {\scriptsize $\pm$6.7} \\
DP-Dfs & 63.6 & \textcolor{gray}{37.3} & \textcolor{gray}{35.8} & \textcolor{gray}{40.9} & \textcolor{gray}{43.6} & 66.5 & \textcolor{gray}{41.5} & \textcolor{gray}{40.7} & \textcolor{gray}{28.1} & \textcolor{gray}{39.9} & 68.4 & \textcolor{gray}{47.5} & \textcolor{gray}{42.7} & \textcolor{gray}{36.2} & \textcolor{gray}{37.3} \\[-3pt]
 & {\scriptsize $\pm$2.8} & {\scriptsize $\pm$1.6} & {\scriptsize $\pm$8.1} & {\scriptsize $\pm$15.1} & {\scriptsize $\pm$17.0} & {\scriptsize $\pm$3.6} & {\scriptsize $\pm$71.9} & {\scriptsize $\pm$48.8} & {\scriptsize $\pm$51.9} & {\scriptsize $\pm$59.9} & {\scriptsize $\pm$6.1} & {\scriptsize $\pm$33.6} & {\scriptsize $\pm$32.7} & {\scriptsize $\pm$21.7} & {\scriptsize $\pm$8.1} \\
EPSVec & \textcolor{gray}{51.3} & \textcolor{gray}{45.9} & 55.9 & 58.0 & 52.7 & \textcolor{gray}{49.8} & \textcolor{gray}{51.6} & 58.9 & 59.4 & 54.7 & 53.0 & \textcolor{gray}{40.9} & \textbf{57.3} & \textbf{57.6} & 52.1 \\[-3pt]
 & {\scriptsize $\pm$2.7} & {\scriptsize $\pm$7.7} & {\scriptsize $\pm$3.4} & {\scriptsize $\pm$3.7} & {\scriptsize $\pm$0.7} & {\scriptsize $\pm$3.6} & {\scriptsize $\pm$12.8} & {\scriptsize $\pm$3.9} & {\scriptsize $\pm$0.3} & {\scriptsize $\pm$2.8} & {\scriptsize $\pm$1.6} & {\scriptsize $\pm$3.9} & {\scriptsize $\pm$1.1} & {\scriptsize $\pm$2.2} & {\scriptsize $\pm$2.6} \\
\hline
\multicolumn{16}{|c|}{\textsc{Dataset ZsTfns} (\textbf{Random}: 48.7, \textbf{Majority}: 65.6)} \\ \hline
\cellcolor{gray!20} & \multicolumn{5}{c|}{\cellcolor{gray!20}\textbf{ProsusAI-finbert (original=88.9{\scriptsize$\pm$0.7})}} & \multicolumn{5}{c|}{\cellcolor{gray!20}\textbf{finbert-tone (original=87.6{\scriptsize$\pm$1.3})}} & \multicolumn{5}{c|}{\cellcolor{gray!20}\textbf{FinancialBERT (original=84.3{\scriptsize$\pm$1.1})}} \\
Method & $\epsilon=\infty$ & $\epsilon=0.5$ & $\epsilon=1$ & $\epsilon=2$ & $\epsilon=4$ & $\epsilon=\infty$ & $\epsilon=0.5$ & $\epsilon=1$ & $\epsilon=2$ & $\epsilon=4$ & $\epsilon=\infty$ & $\epsilon=0.5$ & $\epsilon=1$ & $\epsilon=2$ & $\epsilon=4$ \\
DP-Gen & \textbf{86.6} & \textcolor{gray}{62.5} & \textcolor{gray}{64.9} & \textcolor{gray}{65.3} & \textbf{71.0} & \textbf{85.9} & \textcolor{gray}{62.1} & \textbf{68.0} & \textcolor{gray}{67.2} & \textcolor{gray}{67.2} & \textbf{82.8} & \textcolor{gray}{65.2} & \textcolor{gray}{67.0} & \textcolor{gray}{63.4} & \textbf{68.5} \\[-3pt]
 & {\scriptsize $\pm$0.9} & {\scriptsize $\pm$4.9} & {\scriptsize $\pm$11.1} & {\scriptsize $\pm$5.8} & {\scriptsize $\pm$1.6} & {\scriptsize $\pm$1.2} & {\scriptsize $\pm$1.1} & {\scriptsize $\pm$4.0} & {\scriptsize $\pm$3.0} & {\scriptsize $\pm$9.3} & {\scriptsize $\pm$1.1} & {\scriptsize $\pm$3.5} & {\scriptsize $\pm$7.4} & {\scriptsize $\pm$8.0} & {\scriptsize $\pm$9.2} \\
AUG-PE & 69.6 & \textcolor{gray}{\textbf{64.2}} & \textcolor{gray}{\textbf{65.1}} & \textbf{67.2} & \textcolor{gray}{67.1} & 70.9 & \textcolor{gray}{\textbf{66.4}} & \textcolor{gray}{64.9} & \textbf{67.5} & \textbf{67.8} & 69.5 & \textbf{67.3} & \textbf{69.1} & \textbf{68.2} & \textcolor{gray}{66.6} \\[-3pt]
 & {\scriptsize $\pm$2.6} & {\scriptsize $\pm$4.9} & {\scriptsize $\pm$5.3} & {\scriptsize $\pm$2.7} & {\scriptsize $\pm$6.2} & {\scriptsize $\pm$0.6} & {\scriptsize $\pm$0.6} & {\scriptsize $\pm$1.4} & {\scriptsize $\pm$2.5} & {\scriptsize $\pm$2.0} & {\scriptsize $\pm$7.0} & {\scriptsize $\pm$3.7} & {\scriptsize $\pm$3.1} & {\scriptsize $\pm$3.7} & {\scriptsize $\pm$2.6} \\
DP-Dfs & 74.6 & -- & \textcolor{gray}{49.1} & \textcolor{gray}{45.9} & \textcolor{gray}{39.8} & 74.8 & -- & \textcolor{gray}{65.3} & \textcolor{gray}{55.1} & \textcolor{gray}{57.4} & 74.5 & -- & \textcolor{gray}{62.5} & \textcolor{gray}{62.3} & \textcolor{gray}{63.9} \\[-3pt]
 & {\scriptsize $\pm$4.7} &  & {\scriptsize $\pm$33.5} & {\scriptsize $\pm$10.0} & {\scriptsize $\pm$14.5} & {\scriptsize $\pm$2.7} &  & {\scriptsize $\pm$3.4} & {\scriptsize $\pm$12.0} & {\scriptsize $\pm$6.3} & {\scriptsize $\pm$4.7} &  & {\scriptsize $\pm$7.0} & {\scriptsize $\pm$5.2} & {\scriptsize $\pm$9.2} \\
EPSVec & \textcolor{gray}{52.8} & \textcolor{gray}{35.5} & \textcolor{gray}{60.5} & \textcolor{gray}{49.5} & \textcolor{gray}{53.3} & \textcolor{gray}{56.8} & \textcolor{gray}{40.6} & \textcolor{gray}{62.6} & \textcolor{gray}{52.1} & \textcolor{gray}{59.8} & \textcolor{gray}{61.0} & \textcolor{gray}{53.1} & \textcolor{gray}{62.6} & \textcolor{gray}{57.4} & \textcolor{gray}{53.9} \\[-3pt]
 & {\scriptsize $\pm$6.3} & {\scriptsize $\pm$9.1} & {\scriptsize $\pm$9.6} & {\scriptsize $\pm$3.8} & {\scriptsize $\pm$0.9} & {\scriptsize $\pm$6.5} & {\scriptsize $\pm$10.2} & {\scriptsize $\pm$4.2} & {\scriptsize $\pm$1.9} & {\scriptsize $\pm$1.4} & {\scriptsize $\pm$6.6} & {\scriptsize $\pm$7.3} & {\scriptsize $\pm$1.3} & {\scriptsize $\pm$6.5} & {\scriptsize $\pm$12.1} \\
\hline
\multicolumn{16}{|c|}{\textsc{Dataset Asylax} (\textbf{Random}: 38.4, \textbf{Majority}: 51.4)} \\ \hline
\cellcolor{gray!20} & \multicolumn{5}{c|}{\cellcolor{gray!20}\textbf{legal-bert-base (original=65.6{\scriptsize$\pm$5.0})}} & \multicolumn{5}{c|}{\cellcolor{gray!20}\textbf{InLegalBERT (original=65.3{\scriptsize$\pm$0.7})}} & \multicolumn{5}{c|}{\cellcolor{gray!20}\textbf{Roberta-base (original=58.5{\scriptsize$\pm$2.1})}} \\
Method & $\epsilon=\infty$ & $\epsilon=0.5$ & $\epsilon=1$ & $\epsilon=2$ & $\epsilon=4$ & $\epsilon=\infty$ & $\epsilon=0.5$ & $\epsilon=1$ & $\epsilon=2$ & $\epsilon=4$ & $\epsilon=\infty$ & $\epsilon=0.5$ & $\epsilon=1$ & $\epsilon=2$ & $\epsilon=4$ \\
DP-Gen & \textbf{58.3} & \textcolor{gray}{26.6} & \textcolor{gray}{36.1} & \textcolor{gray}{30.1} & \textcolor{gray}{30.3} & \textbf{57.5} & \textcolor{gray}{27.4} & \textcolor{gray}{47.1} & \textcolor{gray}{23.1} & \textcolor{gray}{33.2} & \textbf{55.9} & \textcolor{gray}{27.6} & \textcolor{gray}{29.7} & \textcolor{gray}{30.4} & \textcolor{gray}{34.0} \\[-3pt]
 & {\scriptsize $\pm$7.1} & {\scriptsize $\pm$5.5} & {\scriptsize $\pm$22.0} & {\scriptsize $\pm$13.9} & {\scriptsize $\pm$3.3} & {\scriptsize $\pm$2.3} & {\scriptsize $\pm$1.0} & {\scriptsize $\pm$9.8} & {\scriptsize $\pm$6.6} & {\scriptsize $\pm$8.9} & {\scriptsize $\pm$3.2} & {\scriptsize $\pm$0.0} & {\scriptsize $\pm$5.9} & {\scriptsize $\pm$6.0} & {\scriptsize $\pm$7.0} \\
AUG-PE & \textcolor{gray}{48.8} & \textcolor{gray}{\textbf{51.4}} & \textcolor{gray}{\textbf{51.3}} & \textcolor{gray}{\textbf{51.4}} & \textcolor{gray}{\textbf{51.0}} & \textcolor{gray}{48.0} & \textcolor{gray}{\textbf{51.4}} & \textcolor{gray}{\textbf{50.2}} & \textcolor{gray}{\textbf{51.3}} & \textcolor{gray}{\textbf{51.4}} & \textcolor{gray}{44.9} & \textcolor{gray}{\textbf{51.4}} & \textcolor{gray}{\textbf{49.5}} & \textcolor{gray}{51.0} & \textcolor{gray}{40.7} \\[-3pt]
 & {\scriptsize $\pm$4.2} & {\scriptsize $\pm$0.0} & {\scriptsize $\pm$0.4} & {\scriptsize $\pm$0.2} & {\scriptsize $\pm$0.6} & {\scriptsize $\pm$7.7} & {\scriptsize $\pm$0.1} & {\scriptsize $\pm$1.4} & {\scriptsize $\pm$0.8} & {\scriptsize $\pm$0.4} & {\scriptsize $\pm$8.4} & {\scriptsize $\pm$0.0} & {\scriptsize $\pm$3.9} & {\scriptsize $\pm$0.5} & {\scriptsize $\pm$9.4} \\
DP-Dfs & \textcolor{gray}{53.7} & \textcolor{gray}{\textbf{51.4}} & \textcolor{gray}{27.7} & \textcolor{gray}{22.5} & \textcolor{gray}{34.8} & 55.2 & \textcolor{gray}{\textbf{51.4}} & \textcolor{gray}{35.0} & \textcolor{gray}{30.2} & \textcolor{gray}{27.1} & \textcolor{gray}{50.2} & \textcolor{gray}{\textbf{51.4}} & \textcolor{gray}{29.6} & \textcolor{gray}{\textbf{51.4}} & \textcolor{gray}{\textbf{51.3}} \\[-3pt]
 & {\scriptsize $\pm$3.4} & {\scriptsize $\pm$0.0} & {\scriptsize $\pm$0.1} & {\scriptsize $\pm$5.2} & {\scriptsize $\pm$13.6} & {\scriptsize $\pm$2.5} & {\scriptsize $\pm$0.0} & {\scriptsize $\pm$28.0} & {\scriptsize $\pm$18.0} & {\scriptsize $\pm$14.9} & {\scriptsize $\pm$4.4} & {\scriptsize $\pm$0.0} & {\scriptsize $\pm$26.7} & {\scriptsize $\pm$0.0} & {\scriptsize $\pm$0.4} \\
EPSVec & \textcolor{gray}{31.4} & \textcolor{gray}{30.7} & \textcolor{gray}{30.0} & \textcolor{gray}{26.2} & \textcolor{gray}{25.2} & \textcolor{gray}{33.0} & \textcolor{gray}{33.6} & \textcolor{gray}{29.9} & \textcolor{gray}{27.1} & \textcolor{gray}{27.6} & \textcolor{gray}{29.8} & \textcolor{gray}{33.8} & \textcolor{gray}{34.2} & \textcolor{gray}{26.6} & \textcolor{gray}{30.2} \\[-3pt]
 & {\scriptsize $\pm$1.3} & {\scriptsize $\pm$5.2} & {\scriptsize $\pm$10.2} & {\scriptsize $\pm$2.5} & {\scriptsize $\pm$6.7} & {\scriptsize $\pm$7.3} & {\scriptsize $\pm$4.9} & {\scriptsize $\pm$10.4} & {\scriptsize $\pm$2.2} & {\scriptsize $\pm$7.5} & {\scriptsize $\pm$7.8} & {\scriptsize $\pm$7.1} & {\scriptsize $\pm$12.6} & {\scriptsize $\pm$6.3} & {\scriptsize $\pm$5.8} \\
\hline
\multicolumn{16}{|c|}{\textsc{Dataset Eurlex} (\textbf{Random}: 24.1, \textbf{Majority}: 36.1)} \\ \hline
\cellcolor{gray!20} & \multicolumn{5}{c|}{\cellcolor{gray!20}\textbf{legal-bert-base (original=89.9{\scriptsize$\pm$0.6})}} & \multicolumn{5}{c|}{\cellcolor{gray!20}\textbf{InLegalBERT (original=89.8{\scriptsize$\pm$0.6})}} & \multicolumn{5}{c|}{\cellcolor{gray!20}\textbf{Roberta-base (original=89.3{\scriptsize$\pm$0.8})}} \\
Method & $\epsilon=\infty$ & $\epsilon=0.5$ & $\epsilon=1$ & $\epsilon=2$ & $\epsilon=4$ & $\epsilon=\infty$ & $\epsilon=0.5$ & $\epsilon=1$ & $\epsilon=2$ & $\epsilon=4$ & $\epsilon=\infty$ & $\epsilon=0.5$ & $\epsilon=1$ & $\epsilon=2$ & $\epsilon=4$ \\
DP-Gen & 63.4 & 49.0 & \textcolor{gray}{28.2} & \textcolor{gray}{37.2} & \textcolor{gray}{35.9} & 60.9 & 49.4 & \textcolor{gray}{33.0} & \textcolor{gray}{38.2} & 43.5 & 59.5 & 52.0 & \textcolor{gray}{38.1} & 39.6 & 46.4 \\[-3pt]
 & {\scriptsize $\pm$3.1} & {\scriptsize $\pm$6.9} & {\scriptsize $\pm$2.3} & {\scriptsize $\pm$0.9} & {\scriptsize $\pm$8.5} & {\scriptsize $\pm$3.2} & {\scriptsize $\pm$1.4} & {\scriptsize $\pm$8.8} & {\scriptsize $\pm$3.6} & {\scriptsize $\pm$8.3} & {\scriptsize $\pm$5.6} & {\scriptsize $\pm$5.2} & {\scriptsize $\pm$16.2} & {\scriptsize $\pm$3.4} & {\scriptsize $\pm$2.8} \\
AUG-PE & 61.1 & \textbf{51.7} & \textbf{52.0} & \textbf{52.9} & \textbf{57.3} & 62.9 & \textbf{54.4} & \textbf{55.2} & \textbf{57.4} & \textbf{55.4} & 63.6 & \textbf{53.8} & \textbf{53.8} & \textbf{59.7} & \textbf{60.1} \\[-3pt]
 & {\scriptsize $\pm$0.8} & {\scriptsize $\pm$7.8} & {\scriptsize $\pm$1.6} & {\scriptsize $\pm$2.1} & {\scriptsize $\pm$1.8} & {\scriptsize $\pm$2.8} & {\scriptsize $\pm$5.4} & {\scriptsize $\pm$1.8} & {\scriptsize $\pm$1.1} & {\scriptsize $\pm$3.2} & {\scriptsize $\pm$1.5} & {\scriptsize $\pm$7.6} & {\scriptsize $\pm$1.8} & {\scriptsize $\pm$4.8} & {\scriptsize $\pm$4.5} \\
DP-Dfs & \textbf{71.4} & -- & \textcolor{gray}{31.7} & 39.1 & 42.4 & \textbf{70.3} & -- & \textcolor{gray}{30.6} & \textcolor{gray}{33.5} & 38.6 & \textbf{65.5} & -- & \textcolor{gray}{30.4} & \textcolor{gray}{34.2} & \textcolor{gray}{35.2} \\[-3pt]
 & {\scriptsize $\pm$4.5} &  & {\scriptsize $\pm$0.7} & {\scriptsize $\pm$10.1} & {\scriptsize $\pm$9.9} & {\scriptsize $\pm$3.7} &  & {\scriptsize $\pm$3.1} & {\scriptsize $\pm$4.8} & {\scriptsize $\pm$8.0} & {\scriptsize $\pm$8.4} &  & {\scriptsize $\pm$6.3} & {\scriptsize $\pm$2.6} & {\scriptsize $\pm$4.4} \\
EPSVec & 55.4 & \textcolor{gray}{32.7} & 38.3 & \textcolor{gray}{33.6} & \textcolor{gray}{27.6} & 57.6 & 39.7 & 42.3 & 40.6 & \textcolor{gray}{36.3} & 56.8 & \textcolor{gray}{35.1} & 51.1 & 45.0 & \textcolor{gray}{29.0} \\[-3pt]
 & {\scriptsize $\pm$6.9} & {\scriptsize $\pm$6.1} & {\scriptsize $\pm$12.1} & {\scriptsize $\pm$6.4} & {\scriptsize $\pm$9.7} & {\scriptsize $\pm$4.5} & {\scriptsize $\pm$5.1} & {\scriptsize $\pm$10.8} & {\scriptsize $\pm$10.1} & {\scriptsize $\pm$3.7} & {\scriptsize $\pm$2.9} & {\scriptsize $\pm$2.9} & {\scriptsize $\pm$10.0} & {\scriptsize $\pm$12.5} & {\scriptsize $\pm$8.1} \\
\hline
\end{tabular}}
\caption{As Table~\ref{tab:utility-llama-bio}, for the finance and legal datasets.}
\label{tab:utility-llama-finlegal}
\end{table*}

\textbf{Utility}
Notably, even without privacy constraints ($\varepsilon=\infty$), synthetic data rarely matches real-data baselines, suggesting that \textbf{the evaluated generators do not fully capture the domain-specific distributions required for downstream learning.}
Only in a few cases do DP-Gen synthetic data, without privacy constraints, enable downstream models to match real-data performance (e.g., \textsc{N2C2'08} on Clinical-LongFormer, \textsc{ZsTfns} on all models, \textsc{AsyLax} on Roberta-base). 
AUG-PE lags behind DP-Gen without a privacy budget on all datasets except, for most classifiers, \textsc{Eurlex}, but is frequently the strongest method
under DP (e.g., \textsc{PsyTAR} at $\varepsilon \geq 1$ and \textsc{Eurlex}).
\textbf{No method dominates under DP.} The best method varies by dataset: DP-Gen on \textsc{Lc37Fnd}, AUG-PE on \textsc{PsyTAR} and \textsc{Eurlex},
DP-Dfs on \textsc{N2C2'08}, and EPSVec on \textsc{HoC}. 
DP-Dfs is the most volatile: it is the strongest method in several settings, yet under strong
privacy constraints ($\varepsilon=0.5$ for \textsc{PsyTAR}, \textsc{ZsTfns},
and \textsc{Eurlex}; $\varepsilon \in \{0.5, 1\}$ for \textsc{DMSAFN}) it
collapses to degenerate outputs such as empty or random strings.
Grey cells in Tables~\ref{tab:utility-llama-bio} and~\ref{tab:utility-llama-finlegal} mark scores not significantly above the majority baseline; on \textsc{AsyLax}, this holds for every method under DP.
\textbf{Utility alone does not capture generation quality:} DP-Dfs attains
the highest downstream score on \textsc{Eurlex} at $\varepsilon=\infty$, yet
its fidelity is inferior to DP-Gen, highlighting the need to assess both
utility and distributional fidelity.

For most datasets and downstream models, there is no consistently clear relationship between the privacy budget and performance:  \textbf{stricter privacy constraints do not always correspond to decreased performance}. 
This observation may be attributed to two factors: first, higher privacy bounds are not being fully utilised for better utility; second, it is possible that strong privacy protections do not necessarily require compromising the utility of synthetic data for downstream tasks. Given the complexity of the data, the under-utilization of higher privacy budgets is likely the primary factor.

After discounting random/majority baselines and measuring improvement over the stronger baseline, rescaled by the gap between that baseline and real-data performance, the improvement ranges without privacy guarantees are $[43.88\%, 92.27\%]$, $[-24.46\%, 64.52\%]$, $[-63.16\%, 65.61\%]$, and $[-140.85\%, 48.92\%]$ for DP-Gen, AUG-PE, DP-Dfs, and EPSVec, respectively, where negative values indicate performance below the baseline. Under DP ($\varepsilon \leq 4$), the maximum observed improvements are  $52.91\%/66.03\%/62.86\%/76.10\%$, showing that while DP methods can still outperform naive  baselines, all fall well short of real-data performance.

Performance is highly dataset-dependent: under DP, the best result of DP-Gen ranges from $-30.94\%$ on \textsc{AsyLax} to $52.91\%$ on \textsc{MIMIC}; AUG-PE from $-1.37\%$ on \textsc{HoC} to $66.03\%$ on \textsc{MIMIC}; DP-Dfs from $-38.46\%$ on \textsc{PsyTAR} to $62.86\%$ on \textsc{DMSAFN}; and EPSVec from $-128.06\%$ on \textsc{AsyLax} to $76.10\%$ on \textsc{DMSAFN}. No dataset is uniformly easy or hard across methods, underscoring that the effectiveness of DP text generation is strongly domain- and dataset-specific.

\begin{table*}[t]
\centering
\scriptsize
\setlength{\tabcolsep}{2.5pt}
\renewcommand{\arraystretch}{0.4}
\begin{tabular}{l|r@{\hspace{3pt}/\hspace{3pt}}r@{\hspace{3pt}/\hspace{3pt}}l|r@{\hspace{3pt}/\hspace{3pt}}r@{\hspace{3pt}/\hspace{3pt}}l|r@{\hspace{3pt}/\hspace{3pt}}r@{\hspace{3pt}/\hspace{3pt}}l|r@{\hspace{3pt}/\hspace{3pt}}r@{\hspace{3pt}/\hspace{3pt}}l|r@{\hspace{3pt}/\hspace{3pt}}r@{\hspace{3pt}/\hspace{3pt}}l}
\toprule
\textbf{Method} & \multicolumn{3}{c|}{$\epsilon=\infty$} & \multicolumn{3}{c|}{$\epsilon=4$} & \multicolumn{3}{c|}{$\epsilon=2$} & \multicolumn{3}{c|}{$\epsilon=1$} & \multicolumn{3}{c}{$\epsilon=0.5$} \\
 & $\mathcal{M}\uparrow$ & $\mathcal{N}\downarrow$ & $\mathcal{L}\downarrow$ & $\mathcal{M}\uparrow$ & $\mathcal{N}\downarrow$ & $\mathcal{L}\downarrow$ & $\mathcal{M}\uparrow$ & $\mathcal{N}\downarrow$ & $\mathcal{L}\downarrow$ & $\mathcal{M}\uparrow$ & $\mathcal{N}\downarrow$ & $\mathcal{L}\downarrow$ & $\mathcal{M}\uparrow$ & $\mathcal{N}\downarrow$ & $\mathcal{L}\downarrow$ \\
\midrule
\rowcolor{gray!20} \textsc{HoC} & \multicolumn{15}{c}{(\textbf{Original}: 0.99 / 2.19 / 0.004)} \\
DP-Gen & \textbf{0.65} & 3.71 & \textbf{0.06} & \textbf{0.18} & 5.26 & 0.40 & \textbf{0.16} & 5.23 & 0.417 & \textbf{0.14} & 5.18 & 0.44 & \textbf{0.12} & 5.21 & \textbf{0.47} \\
AUG-PE & 0.01 & 3.80 & 1.49 & 0.01 & 3.92 & 1.40 & 0.01 & \textbf{4.06} & 1.35 & 0.01 & 4.45 & 1.25 & 0.01 & 4.72 & 1.25 \\
DP-Dfs & 0.01 & 4.70 & \textbf{0.06} & 0.01 & 6.49 & \textbf{0.36} & 0.01 & 6.62 & \textbf{0.37} & 0.01 & 6.855 & \textbf{0.42} & 0.01 & 6.972 & 1.04 \\
EPSVec & 0.01 & \textbf{3.38} & 2.74 & 0.01 & \textbf{3.76} & 2.85 & 0.01 & 4.48 & 3.56 & 0.01 & \textbf{3.39} & 2.78 & 0.01 & \textbf{4.13} & 3.13 \\
\midrule
\rowcolor{gray!20} \textsc{N2C2'08} & \multicolumn{15}{c}{(\textbf{Original}: 0.99 / 0.90 / 0.17)} \\
DP-Gen & \textbf{0.42} & \textbf{2.34} & \textbf{0.75} & 0.02 & 9.07 & \textbf{1.37} & \textbf{0.02} & 6.79 & \textbf{1.40} & \textbf{0.02} & 7.12 & \textbf{1.42} & \textbf{0.02} & 7.19 & \textbf{1.54} \\
AUG-PE & 0.02 & 5.75 & 3.72 & \textbf{0.03} & 6.50 & 3.80 & \textbf{0.02} & 6.66 & 3.87 & \textbf{0.02} & 6.75 & 3.63 & \textbf{0.02} & 6.98 & 3.42 \\
DP-Dfs & 0.01 & 4.97 & 3.09 & 0.01 & 6.06 & 2.91 & 0.01 & 6.31 & 2.91 & 0.01 & 6.45 & 2.91 & 0.01 & 6.719 & 4.01 \\
EPSVec & 0.02 & 7.17 & 1.99 & 0.02 & \textbf{5.83} & 2.08 & \textbf{0.02} & \textbf{5.36} & 2.30 & \textbf{0.02} & \textbf{4.72} & 2.70 & \textbf{0.02} & \textbf{4.23} & 2.98 \\
\midrule
\rowcolor{gray!20} \textsc{PsyTAR} & \multicolumn{15}{c}{(\textbf{Original}: 0.99 / 1.08 / 0.007)} \\
DP-Gen & \textbf{0.61} & 2.60 & \textbf{0.03} & \textbf{0.42} & 3.17 & \textbf{0.03} & \textbf{0.34} & \textbf{3.24} & \textbf{0.05} & \textbf{0.37} & 3.63 & \textbf{0.03} & \textbf{0.35} & 3.71 & \textbf{0.03} \\
AUG-PE & 0.02 & 4.38 & 3.21 & 0.02 & 4.62 & 3.27 & 0.02 & 4.87 & 3.23 & 0.02 & 5.24 & 3.34 & 0.02 & 5.49 & 3.38 \\
DP-Dfs & 0.06 & 2.62 & 0.12 & 0.01 & 8.08 & 0.16 & 0.01 & 8.28 & 0.06 & 0.01 & 8.59 & 0.07 & -- & -- & -- \\
EPSVec & 0.08 & \textbf{2.42} & 6.24 & 0.07 & \textbf{2.37} & 6.33 & 0.04 & 3.56 & 5.48 & 0.08 & \textbf{2.89} & 6.31 & 0.08 & \textbf{2.76} & 7.47 \\
\midrule
\rowcolor{gray!20} \textsc{Mimic} & \multicolumn{15}{c}{(\textbf{Original}: 0.99 / 0.24 / 0.003)} \\
DP-Gen & \textbf{0.96} & \textbf{0.30} & \textbf{0.01} & 0.01 & 3.35 & \textbf{2.69} & 0.01 & 3.65 & 2.59 & 0.01 & 3.80 & 2.49 & 0.01 & 4.89 & \textbf{2.76} \\
AUG-PE & 0.01 & 5.44 & 3.83 & 0.01 & 5.40 & 6.21 & 0.01 & 5.41 & \textbf{2.41} & 0.01 & 5.51 & \textbf{2.42} & 0.01 & 5.79 & 6.20 \\
DP-Dfs & 0.01 & 4.46 & 5.79 & 0.01 & 5.72 & 5.79 & 0.01 & 5.66 & 5.38 & 0.01 & 5.55 & 5.38 & 0.01 & 5.73 & 5.38 \\
EPSVec & 0.01 & 3.03 & 4.92 & 0.01 & \textbf{3.23} & 5.06 & 0.01 & \textbf{3.23} & 5.33 & 0.01 & \textbf{3.18} & 5.34 & 0.01 & \textbf{3.31} & 5.26 \\
\midrule
\rowcolor{gray!20} \textsc{DMSAFN} & \multicolumn{15}{c}{(\textbf{Original}: 0.86 / 1.23 / 0.012)} \\
DP-Gen & \textbf{0.21} & \textbf{2.40} & \textbf{0.06} & \textbf{0.18} & \textbf{2.84} & \textbf{0.02} & \textbf{0.16} & \textbf{2.74} & \textbf{0.04} & \textbf{0.15} & \textbf{3.04} & \textbf{0.03} & \textbf{0.12} & \textbf{3.04} & \textbf{0.06} \\
AUG-PE & 0.01 & 3.85 & 2.36 & 0.01 & 4.41 & 2.35 & 0.01 & 4.66 & 2.34 & 0.01 & 5.17 & 2.24 & 0.01 & 4.63 & 2.30 \\
DP-Dfs & 0.01 & 10.04 & 0.12 & 0.01 & 9.53 & 0.43 & 0.01 & 9.49 & 0.37 & -- & -- & -- & -- & -- & -- \\
EPSVec & 0.01 & 5.62 & 6.47 & 0.01 & 5.12 & 5.78 & 0.01 & 4.89 & 6.01 & 0.01 & 5.19 & 6.04 & 0.01 & 5.30 & 6.85 \\
\midrule
\rowcolor{gray!20} \textsc{Lc37Fnd} & \multicolumn{15}{c}{(\textbf{Original}: 0.99 / 0.88 / 0.002)} \\
DP-Gen & \textbf{0.90} & \textbf{1.17} & \textbf{0.03} & 0.01 & 4.61 & 3.71 & 0.01 & 4.77 & 4.04 & 0.01 & 4.77 & 3.95 & 0.01 & 4.64 & 4.88 \\
AUG-PE & 0.01 & 4.22 & 2.52 & 0.01 & 4.45 & 2.67 & 0.01 & 5.20 & 2.82 & 0.01 & 5.14 & 2.91 & 0.01 & 5.45 & 2.87 \\
DP-Dfs & 0.01 & 7.00 & 1.37 & 0.01 & 6.87 & 1.86 & 0.01 & 7.10 & 1.87 & 0.01 & 7.39 & 1.72 & 0.01 & 7.39 & \textbf{1.65} \\
EPSVec & 0.04 & 2.26 & 1.46 & \textbf{0.03} & \textbf{2.29} & \textbf{1.49} & \textbf{0.04} & \textbf{2.27} & \textbf{1.35} & \textbf{0.04} & \textbf{2.24} & \textbf{1.26} & \textbf{0.02} & \textbf{2.90} & 2.30 \\
\midrule
\rowcolor{gray!20} \textsc{ZsTfns} & \multicolumn{15}{c}{(\textbf{Original}: 0.99 / 1.29 / 0.002)} \\
DP-Gen & \textbf{0.96} & \textbf{1.75} & \textbf{0.01} & \textbf{0.04} & 4.89 & \textbf{0.09} & \textbf{0.04} & 5.04 & \textbf{0.15} & \textbf{0.05} & 5.06 & \textbf{0.19} & \textbf{0.43} & \textbf{2.44} & \textbf{0.01} \\
AUG-PE & 0.02 & 3.55 & 4.36 & 0.02 & 3.78 & 4.39 & 0.02 & \textbf{3.97} & 4.38 & 0.02 & 4.13 & 4.46 & 0.01 & 4.26 & 4.39 \\
DP-Dfs & 0.01 & 7.38 & 0.87 & 0.01 & 7.10 & 0.87 & 0.01 & 7.22 & 0.77 & 0.01 & 7.18 & 1.22 & -- & -- & -- \\
EPSVec & 0.03 & 3.78 & 3.89 & 0.03 & \textbf{3.68} & 3.99 & 0.02 & 4.45 & 4.44 & 0.02 & \textbf{3.88} & 3.84 & 0.02 & 4.09 & 4.64 \\
\midrule
\rowcolor{gray!20} \textsc{AsyLax} & \multicolumn{15}{c}{(\textbf{Original}: 0.98 / 0.11 / 0.004)} \\
DP-Gen & \textbf{0.03} & \textbf{0.61} & \textbf{0.43} & 0.01 & \textbf{3.02} & 6.41 & 0.01 & \textbf{3.24} & 5.95 & 0.01 & 6.20 & \textbf{3.58} & 0.01 & 6.54 & \textbf{1.23} \\
AUG-PE & 0.01 & 5.53 & 1.41 & 0.01 & 4.95 & \textbf{1.41} & 0.01 & 5.55 & \textbf{1.92} & 0.01 & 5.63 & 6.12 & 0.01 & 5.92 & 1.41 \\
DP-Dfs & 0.01 & 8.01 & 3.42 & 0.01 & 8.03 & 3.22 & 0.01 & 8.04 & 3.23 & 0.01 & 7.93 & 4.43 & 0.01 & 8.21 & 6.91 \\
EPSVec & 0.01 & 4.08 & 4.06 & 0.01 & 4.06 & 4.08 & 0.01 & 4.02 & 4.12 & 0.01 & \textbf{4.01} & 4.44 & 0.01 & \textbf{4.01} & 4.14 \\
\midrule
\rowcolor{gray!20} \textsc{Eurlex} & \multicolumn{15}{c}{(\textbf{Original}: 0.98 / 0.26 / 0.006)} \\
DP-Gen & \textbf{0.05} & \textbf{0.52} & 0.06 & \textbf{0.10} & \textbf{0.79} & 0.23 & \textbf{0.07} & \textbf{0.83} & 0.27 & \textbf{0.03} & \textbf{0.93} & 0.22 & 0.01 & \textbf{1.42} & 0.45 \\
AUG-PE & 0.01 & 1.97 & \textbf{0.05} & 0.01 & 2.04 & \textbf{0.06} & 0.01 & 2.12 & \textbf{0.04} & 0.01 & 2.36 & 0.08 & 0.01 & 2.40 & 0.08 \\
DP-Dfs & 0.01 & 9.26 & 0.28 & 0.01 & 8.20 & 0.47 & 0.01 & 8.61 & 0.47 & 0.01 & 8.37 & 0.39 & -- & -- & -- \\
EPSVec & 0.01 & 1.15 & 0.13 & 0.01 & 1.26 & 0.09 & 0.01 & 1.43 & 0.06 & 0.01 & 1.39 & \textbf{0.03} & 0.01 & 1.47 & \textbf{0.04} \\
\bottomrule
\end{tabular}
\caption{Fidelity of Llama-3.2-1B synthetic data to the real test set: MAUVE ($\mathcal{M}$), KL divergence of NER-entity 1-gram distributions ($\mathcal{N}$) and of text-length distributions ($\mathcal{L}$). Bold: the best method per dataset, $\epsilon$ and metric (none when all methods tie). ``--'': not available (DP-Dfs settings whose synthetic data were unusable).}
\label{tab:fidelity-llama}
\end{table*}

\textbf{Fidelity}
We find that \textbf{the fidelity of synthetic data deteriorates under differential privacy constraints}, highlighting the persistent challenge of generating high-quality, domain-specific data.
As shown in Table~\ref{tab:fidelity-llama}, synthetic data diverges from real data more than real train and test splits do on all three metrics, for every method and privacy budget: MAUVE is lower, and the KL divergences of NER-entity and text-length distributions are higher.
Without privacy constraints, DP-Gen can approach the real distribution (e.g., MAUVE of $0.96$ on \textsc{MIMIC} and \textsc{ZsTfns}), but under DP its MAUVE drops close to zero on most datasets and both divergences increase. 
The other methods already diverge substantially at $\varepsilon=\infty$ (MAUVE $\leq 0.08$), and their fidelity changes little once DP is applied.

Relying on the post-processing property of DP, we can recover some utility and fidelity by over-generating synthetic data and filtering it with quality metrics that do not reference the private data (see Appendix~\ref{app:post-processing}); however, the resulting data still
lags behind data generated without privacy constraints.
Collectively, these findings reinforce our finding: \textbf{generating high-quality domain-specific data under differential privacy constraints remains an unresolved challenge}, with fidelity and utility degrading markedly as domain complexity increases, particularly for domains that are less represented in pre-training corpora.

\textbf{Training Data Leakage.} Figure~\ref{img:leakage_corr} suggests that \textbf{if an LLM’s pre-training data overlaps with records from a nominally ``private'' dataset, the model may achieve higher measured synthetic-data quality on that dataset (even under DP), consistent with pre-training exposure influencing downstream generation performance}. To quantify this effect, we measure exact overlap between each private training set and a public web-scale reference corpus using the Infini-gram API, and we use RedPajama as a proxy for LLaMA-style pre-training data.\footnote{Because the most recent LLaMA pre-training corpora are not public, our leakage estimates are necessarily with respect to RedPajama; nevertheless, they provide actionable signals for studying contamination effects.} For each dataset, we compute the maximum MAUVE across privacy levels ($\varepsilon \in \{4, 2, 1, 0.5\}$) and correlate it with the fraction of leaked samples, finding a positive association (Spearman $\rho = 0.76$, $p = 0.04$). This is also consistent with the large gains of AUG-PE on \textsc{PsyTAR}, which, despite gated access, shows 9.5\% estimated overlap.

Because this cross-dataset correlation may be confounded by differences between datasets, we further run a fully controlled experiment with \textsc{OLMo-2}, where pre-training corpora are public. For \textsc{HoC}, where $\approx$50\% of training samples appear in the pre-training corpus, construct private datasets with progressively more contaminated samples to synthesise data from. All downstream classifiers gain performance as contamination increases, with up to 3.4\% utility gain (See Appendix~\ref{app:olmo} for details).
Finally, we find that datasets with higher estimated overlap also tend to show higher membership inference attack performance.

\begin{figure*}[h!]
\includegraphics[width=1\linewidth]{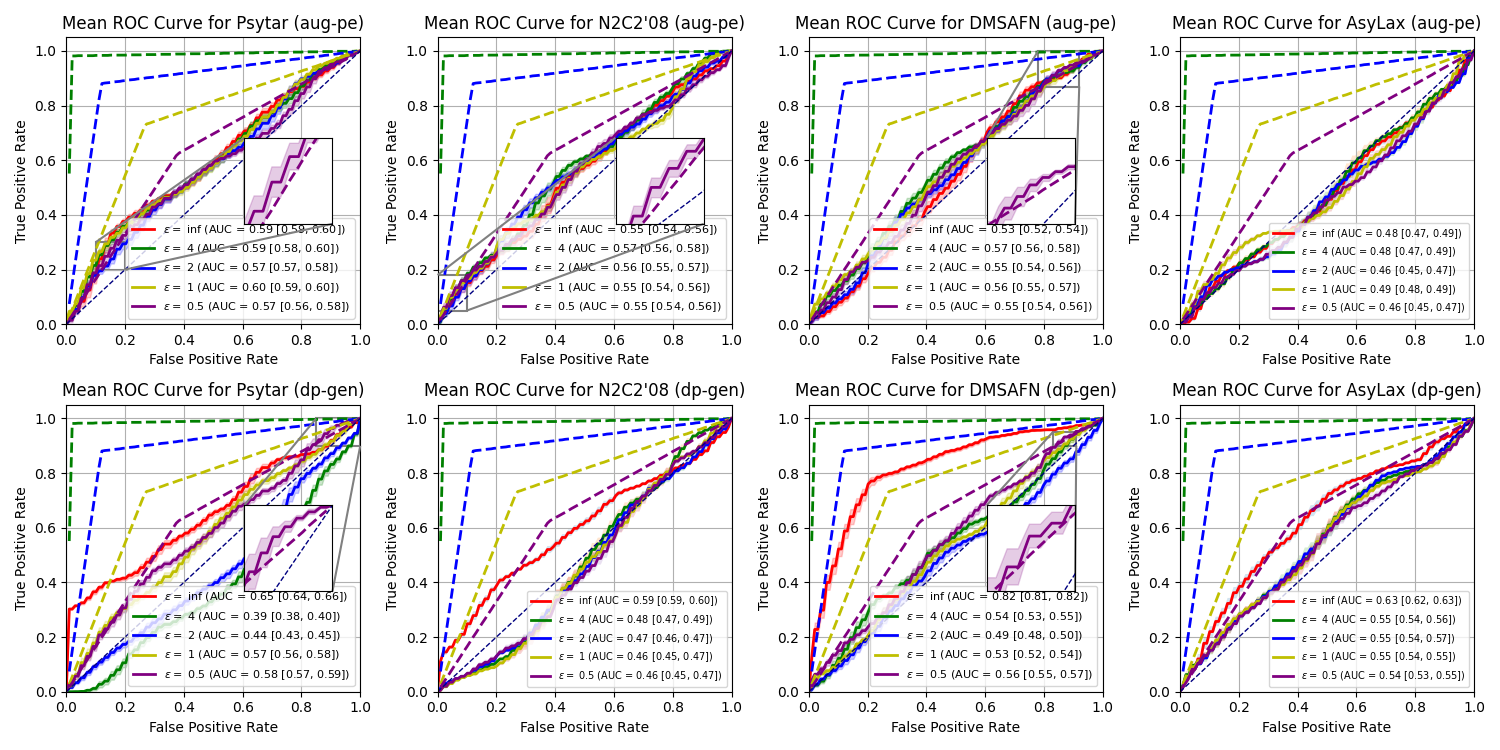}
\caption{
The ROC curves illustrate the performance of membership inference attacks (MIA) on synthetic data generated by AUG-PE and DP-gen. The colourful dashed lines represent the theoretical upper bounds for each corresponding privacy parameter $\epsilon$ from~\cite{Kairouz2015ThePrivacy}. Zoomed sections highlight FPR ranges where the empirical ROC curve lies above the theoretical bound. 
}
\label{img:mia_roc}
\vspace{-1.0em}
\end{figure*}

\textbf{Privacy}
For AUG-PE, as illustrated in Figure~\ref{img:mia_roc}, the attack shows measurable membership signal on \textsc{PsyTAR}, \textsc{n2c2'08} and \textsc{DMSAFN} even under a tight privacy budget ($\varepsilon=0.5$). In some FPR ranges---$[0.1,0.2]$ for \textsc{PsyTAR} and \textsc{n2c2'08}, and $[0.8,0.9]$ for \textsc{DMSAFN}---the mean ROC curve lies above the theoretical bound despite low overall AUC, in line with ~\cite{Carlini2022MembershipPrinciples}'s observation that average attack success may not reflect worst-case leakage (see Appendix~\ref{app:empirical-privacy} for a stricter analysis). 
DP-Gen offers stronger overall protection on \textsc{PsyTAR} and \textsc{N2C2'08}, while exceedances appear at high FPR ($[0.85,1.0]$ for \textsc{PsyTAR}, $[0.8,0.9]$ for \textsc{DMSAFN}).
Two factors may explain why membership signal concentrates at specific operating points. First, our targets are distributional outliers, which represent a worst case relative to normal records: signal concentrated on a few such targets raises attack success at specific operating points without substantially increasing overall AUC. 
Second, \textsc{PsyTAR}, where signal appears for both methods, has relatively high
estimated pre-training overlap and shows large AUG-PE utility gains; one hypothesis is that private fine-tuning more readily reinforces records the generator has already seen during pre-training. The exceedances themselves occur in narrow FPR ranges, several at high FPR, where empirical $\varepsilon$ is estimated from few errors and is sensitive
to sampling variability. Disentangling these factors, e.g., with uncontaminated canaries or a larger number of audit trials, is a promising direction for future work.

\paragraph{Pre-training exposure lies outside the DP guarantee.}
As noted by~\citet{Tramer2024Position:Pretraining}, the DP guarantee is
conditional on the pre-trained generator, so end-to-end privacy risk can
arise from stages outside the DP mechanism, most notably pre-training on
``public'' corpora that overlap with private data. Model-based attacks
(RMIA, LiRA, and a logits-based attack; Appendix~\ref{app:mia_tables})
on downstream classifiers further show that membership signal depends on
the pipeline: on \textsc{PsyTAR}, a classifier trained on DP-synthetic
data is \emph{more} vulnerable than one trained directly on the private
data with DP at the same $\varepsilon$. Such pipeline-dependent behaviour
cannot be read off the nominal $(\varepsilon,\delta)$ budget, motivating
standardised, multi-attack audits of DP synthetic text generation.

\textbf{The mapping from the nominal privacy budget $\epsilon$ to empirical MIA leakage can be non-monotonic.}
As expected, removing DP constraints substantially increases vulnerability (e.g., for \textsc{DP-Gen} with $\epsilon=\infty$, MIA reaches AUC $0.82$ on \textsc{DMSAFN}). However, among DP settings, tighter budgets do not always reduce leakage; for example on \textsc{PsyTAR}, AUC increases from $0.39$ at $\varepsilon=4$ to $0.58$ at $\varepsilon=0.5$, although part of this variation may reflect sampling variability. These observations suggest that $\varepsilon$ alone is an incomplete summary of empirical leakage: by post-processing, $(\varepsilon,\delta)$ upper-bounds leakage from the released synthetic text, but the leakage actually realized below that bound can vary substantially across datasets and pipelines, motivating empirical audits.

\begin{figure}[htbp]
  \centering
  \begin{minipage}{0.52\textwidth}
    \centering
    \includegraphics[width=\linewidth]{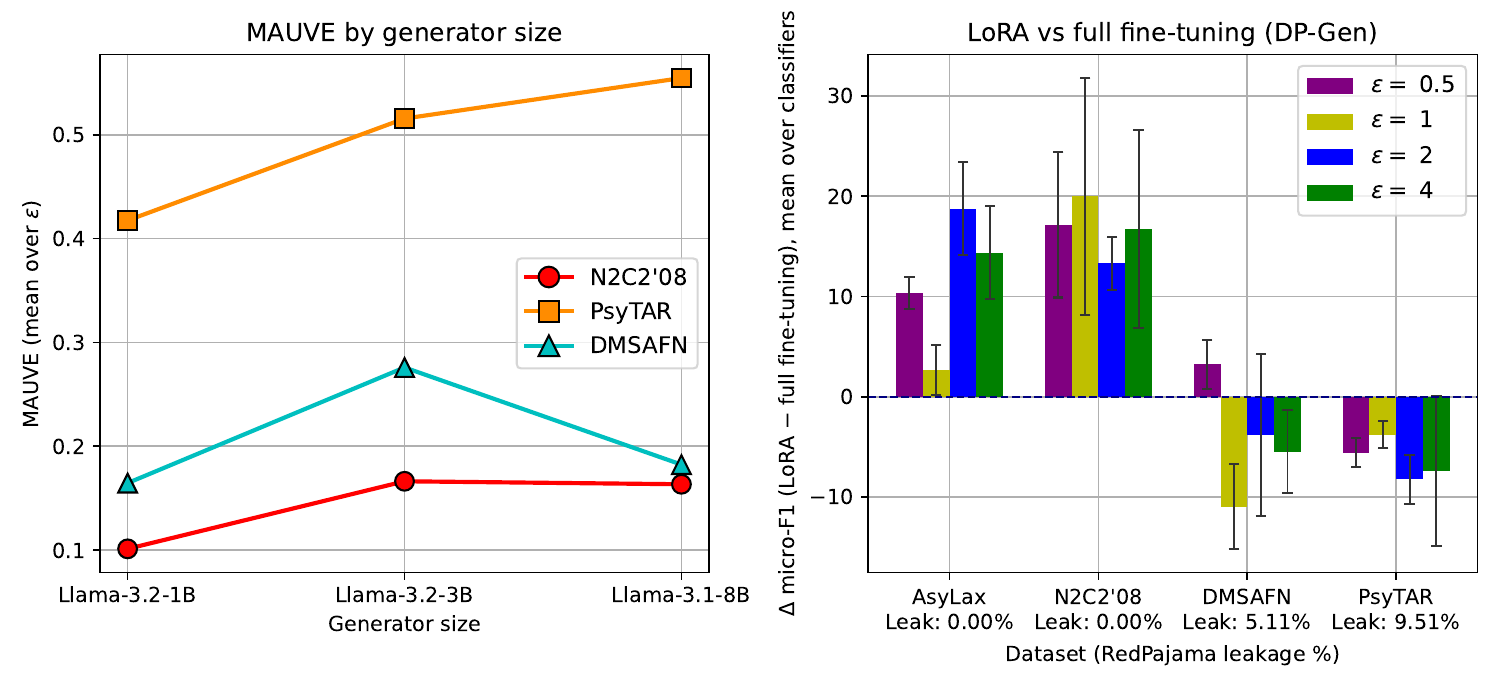}
    \caption{\textbf{Left}: MAUVE (mean over $\varepsilon$) vs.\ generator size. \textbf{Right}: F1 gain of LoRA over full fine-tuning, sorted by estimated pre-training leakage.}
    \label{img:fine-tune-full-lora}
  \end{minipage}
  \hfill
  \begin{minipage}{0.46\textwidth}
    \centering
    \resizebox{\linewidth}{!}{\begin{tabular}{llccccc}
\toprule
 & Size & $\varepsilon=\infty$ & $\varepsilon=4$ & $\varepsilon=2$ & $\varepsilon=1$ & $\varepsilon=0.5$ \\
\midrule
\multirow{3}{*}{\textsc{PsyTAR}}
 & 1B & 73.2\pmci{0.2} & 33.9\pmci{8.4} & 32.8\pmci{2.5} & 34.4\pmci{1.6} & 31.3\pmci{0.4} \\
 & 3B & 74.3\pmci{0.7} & 40.4\pmci{1.5} & 36.1\pmci{0.4} & 31.2\pmci{1.0} & 23.1\pmci{2.3} \\
 & 8B & \textbf{75.7}\pmci{1.1} & \textbf{41.6}\pmci{0.7} & \textbf{41.1}\pmci{0.1} & \textbf{39.8}\pmci{1.5} & \textbf{34.4}\pmci{1.3} \\
\midrule
\multirow{3}{*}{\textsc{N2C2'08}}
 & 1B & 58.2\pmci{7.3} & 67.8\pmci{3.6} & 67.8\pmci{3.1} & \textbf{70.9}\pmci{1.0} & \textbf{70.1}\pmci{3.0} \\
 & 3B & 64.8\pmci{2.5} & 67.4\pmci{10.2} & \textbf{71.0}\pmci{5.5} & 68.4\pmci{3.5} & 69.8\pmci{1.9} \\
 & 8B & \textbf{74.8}\pmci{5.5} & \textbf{69.0}\pmci{1.1} & 69.6\pmci{2.0} & 68.9\pmci{4.4} & 66.6\pmci{8.9} \\
\midrule
\multirow{3}{*}{\textsc{DMSAFN}}
 & 1B & 76.4\pmci{4.0} & \textbf{42.2}\pmci{5.1} & 41.4\pmci{3.2} & 41.5\pmci{3.1} & 47.9\pmci{2.5} \\
 & 3B & \textbf{78.9}\pmci{3.0} & 40.6\pmci{6.5} & \textbf{43.3}\pmci{3.2} & \textbf{42.0}\pmci{0.9} & \textbf{62.6}\pmci{11.0} \\
 & 8B & 78.8\pmci{4.5} & 22.1\pmci{7.2} & 30.1\pmci{5.4} & 41.0\pmci{4.6} & 47.7\pmci{2.8} \\
\midrule
\multirow{3}{*}{\textsc{AsyLax}}
 & 1B & 30.8\pmci{8.7} & \textbf{46.8}\pmci{4.2} & \textbf{46.6}\pmci{5.6} & \textbf{40.3}\pmci{3.0} & \textbf{37.6}\pmci{1.3} \\
 & 3B & \textbf{47.1}\pmci{3.5} & 38.3\pmci{4.7} & 43.6\pmci{3.7} & 37.2\pmci{0.6} & 35.6\pmci{11.3} \\
 & 8B & 46.9\pmci{10.0} & 29.6\pmci{3.8} & 27.7\pmci{1.4} & 27.6\pmci{0.0} & 27.6\pmci{0.1} \\
\bottomrule
\end{tabular}

}
    \captionof{table}{Impact on utility of synthetic data generated by DP-finetuned LLMs of increasing sizes at varying $\epsilon$.}
    \label{tab:text-llama-utility}
  \end{minipage}
\end{figure}
\vspace{-1em}

\textbf{Evaluation of Model Size} To assess the influence of model size on synthetic data quality, we compare Llama-3.2-1B, Llama-3.2-3B, and Llama-3.1-8B. Table~\ref{tab:text-llama-utility} and Figure~\ref{img:fine-tune-full-lora} (left) yield two observations.
(1) Without privacy constraints, F1 generally increases with model size, but under DP this trend does not hold consistently across datasets and privacy budgets (e.g., on \textsc{AsyLax}, the 1B model performs best at every $\varepsilon \leq 4$). 
(2) MAUVE increases with model size on \textsc{PsyTAR} ($0.42 \rightarrow 0.55$) and slightly on \textsc{N2C2'08}, but not on \textsc{DMSAFN}. The pronounced gain on \textsc{PsyTAR}, which has the highest estimated leakage, may hint at a link between model size and pre-training exposure, although the evidence is limited to a few datasets.

\textbf{Evaluation of Fune-tuning Strategy}
Figure~\ref{img:fine-tune-full-lora} (right) shows a pattern aligned with estimated pre-training leakage: on datasets with no detected leakage (\textsc{AsyLax} and \textsc{N2C2'08}), LoRA outperforms full fine-tuning at every $\varepsilon$, whereas on datasets with detected leakage (\textsc{DMSAFN} and \textsc{PsyTAR}) it mostly underperforms, with one exception (\textsc{DMSAFN} at $\varepsilon=0.5$). 
One explanation is that LoRA trains far fewer parameters, so the noise added by DP-SGD is spread over a lower-dimensional update, improving the signal-to-noise ratio~\cite{Yu2021DifferentiallyModels}. 
On datasets already seen during pre-training, full fine-tuning may instead benefit from its greater capacity to reactivate memorised content. Given the small number of datasets and the wide error bars, we present this pattern as suggestive.
\section{Conclusions and Limitations}
This work highlights the persistent challenges in generating high-quality, domain-specific synthetic text with differential privacy. Using a standardized evaluation framework across diverse datasets, we show that current methods struggle to jointly optimize utility, fidelity, and privacy, and our membership-inference audits reveal measurable membership signals even at
tight privacy budgets. Because the nominal guarantee is conditional on the pre-trained
generator and does not account for pre-training exposure, we recommend empirical
privacy validation, particularly when evaluation data may overlap with web-scale
pre-training corpora. While our audits can guide safer deployment of DP synthetic data, they may also inform adversaries about where current methods leak.
\paragraph{Limitations.} \emph{Contamination.} For Llama, overlap is
estimated against a proxy corpus (RedPajama), and exact matching
under-detects paraphrased exposure; only for OLMo-2 can it be measured
against the actual corpora. Our cross-dataset analyses do not isolate the
causal effect of exposure, which we leave to future work (e.g., via
uncontaminated canaries). \emph{Scope.} Main results use the Llama family;
we replicate DP-Gen with OLMo-2 (Appendix~\ref{app:olmo}), but other
backbones may behave differently. \emph{Reproducibility.} Several datasets
require data use agreements; we release fixed split indices and, where
licenses permit, cached outputs. \emph{Auditing.} Targeting outliers
yields a worst-case lower bound on leakage rather than typical-record
risk; the full
protocol is also resource-intensive. \emph{Metrics.} MAUVE depends on the
embedding model, and NER divergence on the tagger.

\begin{ack}
The majority of the computational work for this article was performed on
resources provided by Imperial Global Singapore. We are also grateful for
the support provided by Research IT in the form of access to the
Computational Shared Facility at The University of Manchester and the
computational facilities at the Imperial College Research Computing
Service (DOI: \url{https://doi.org/10.14469/hpc/2232}), and some
computations were performed on resources of the National Supercomputing
Centre, Singapore (\url{https://www.nscc.sg}). We also thank the anonymous
NeurIPS reviewers for their feedback that helped us improve the paper
further.

\paragraph{Funding.} This research is part of the IN-CYPHER programme and
is supported by the National Research Foundation, Prime Minister's
Office, Singapore under its Campus for Research Excellence and
Technological Enterprise (CREATE) programme.

\paragraph{Competing interests.} The authors declare no competing
interests.
\end{ack}

\bibliography{neurips_2026}
\bibliographystyle{unsrtnat}

\newpage
\section*{NeurIPS Paper Checklist}

The checklist is designed to encourage best practices for responsible machine learning research, addressing issues of reproducibility, transparency, research ethics, and societal impact. Do not remove the checklist: {\bf The papers not including the checklist will be desk rejected.} The checklist should follow the references and follow the (optional) supplemental material.  The checklist does NOT count towards the page
limit. 

Please read the checklist guidelines carefully for information on how to answer these questions. For each question in the checklist:
\begin{itemize}
    \item You should answer \answerYes{}, \answerNo{}, or \answerNA{}.
    \item \answerNA{} means either that the question is Not Applicable for that particular paper or the relevant information is Not Available.
    \item Please provide a short (1--2 sentence) justification right after your answer (even for \answerNA). 
\end{itemize}

{\bf The checklist answers are an integral part of your paper submission.} They are visible to the reviewers, area chairs, senior area chairs, and ethics reviewers. You will also be asked to include it (after eventual revisions) with the final version of your paper, and its final version will be published with the paper.

The reviewers of your paper will be asked to use the checklist as one of the factors in their evaluation. While \answerYes{} is generally preferable to \answerNo{}, it is perfectly acceptable to answer \answerNo{} provided a proper justification is given (e.g., error bars are not reported because it would be too computationally expensive'' or ``we were unable to find the license for the dataset we used''). In general, answering \answerNo{} or \answerNA{} is not grounds for rejection. While the questions are phrased in a binary way, we acknowledge that the true answer is often more nuanced, so please just use your best judgment and write a justification to elaborate. All supporting evidence can appear either in the main paper or the supplemental material, provided in appendix. If you answer \answerYes{} to a question, in the justification please point to the section(s) where related material for the question can be found.

IMPORTANT, please:
\begin{itemize}
    \item {\bf Delete this instruction block, but keep the section heading ``NeurIPS Paper Checklist"},
    \item  {\bf Keep the checklist subsection headings, questions/answers and guidelines below.}
    \item {\bf Do not modify the questions and only use the provided macros for your answers}.
\end{itemize}


\begin{enumerate}

\item {\bf Claims}
    \item[] Question: Do the main claims made in the abstract and introduction accurately reflect the paper's contributions and scope?
    \item[] Answer: \answerYes{} 
    \item[] Justification: The abstract and introduction clearly state the primary contributions and the scope of the work, which are fully substantiated in the main text and validated by the experimental results.
    \item[] Guidelines:
    \begin{itemize}
        \item The answer \answerNA{} means that the abstract and introduction do not include the claims made in the paper.
        \item The abstract and/or introduction should clearly state the claims made, including the contributions made in the paper and important assumptions and limitations. A \answerNo{} or \answerNA{} answer to this question will not be perceived well by the reviewers. 
        \item The claims made should match theoretical and experimental results, and reflect how much the results can be expected to generalize to other settings. 
        \item It is fine to include aspirational goals as motivation as long as it is clear that these goals are not attained by the paper. 
    \end{itemize}

\item {\bf Limitations}
    \item[] Question: Does the paper discuss the limitations of the work performed by the authors?
    \item[] Answer: \answerYes{} 
    \item[] Justification: The limitations of our approach are explicitly discussed in Section 5, where we address potential weaknesses to ensure transparency and guide future work.
    \item[] Guidelines:
    \begin{itemize}
        \item The answer \answerNA{} means that the paper has no limitation while the answer \answerNo{} means that the paper has limitations, but those are not discussed in the paper. 
        \item The authors are encouraged to create a separate ``Limitations'' section in their paper.
        \item The paper should point out any strong assumptions and how robust the results are to violations of these assumptions (e.g., independence assumptions, noiseless settings, model well-specification, asymptotic approximations only holding locally). The authors should reflect on how these assumptions might be violated in practice and what the implications would be.
        \item The authors should reflect on the scope of the claims made, e.g., if the approach was only tested on a few datasets or with a few runs. In general, empirical results often depend on implicit assumptions, which should be articulated.
        \item The authors should reflect on the factors that influence the performance of the approach. For example, a facial recognition algorithm may perform poorly when image resolution is low or images are taken in low lighting. Or a speech-to-text system might not be used reliably to provide closed captions for online lectures because it fails to handle technical jargon.
        \item The authors should discuss the computational efficiency of the proposed algorithms and how they scale with dataset size.
        \item If applicable, the authors should discuss possible limitations of their approach to address problems of privacy and fairness.
        \item While the authors might fear that complete honesty about limitations might be used by reviewers as grounds for rejection, a worse outcome might be that reviewers discover limitations that aren't acknowledged in the paper. The authors should use their best judgment and recognize that individual actions in favor of transparency play an important role in developing norms that preserve the integrity of the community. Reviewers will be specifically instructed to not penalize honesty concerning limitations.
    \end{itemize}

\item {\bf Theory assumptions and proofs}
    \item[] Question: For each theoretical result, does the paper provide the full set of assumptions and a complete (and correct) proof?
    \item[] Answer: \answerNA{} 
    \item[] Justification: This paper does not propose new theoretical results; instead, it builds upon existing well-established theory, and all relevant assumptions or proofs are referenced from prior work.
    \item[] Guidelines:
    \begin{itemize}
        \item The answer \answerNA{} means that the paper does not include theoretical results. 
        \item All the theorems, formulas, and proofs in the paper should be numbered and cross-referenced.
        \item All assumptions should be clearly stated or referenced in the statement of any theorems.
        \item The proofs can either appear in the main paper or the supplemental material, but if they appear in the supplemental material, the authors are encouraged to provide a short proof sketch to provide intuition. 
        \item Inversely, any informal proof provided in the core of the paper should be complemented by formal proofs provided in appendix or supplemental material.
        \item Theorems and Lemmas that the proof relies upon should be properly referenced. 
    \end{itemize}

    \item {\bf Experimental result reproducibility}
    \item[] Question: Does the paper fully disclose all the information needed to reproduce the main experimental results of the paper to the extent that it affects the main claims and/or conclusions of the paper (regardless of whether the code and data are provided or not)?
    \item[] Answer: \answerYes{} 
    \item[] Justification: All details required to reproduce the main experimental results---including datasets, evaluation metrics, model configurations, and hyperparameters---are provided in Sections 4 and the appendix.
    \item[] Guidelines:
    \begin{itemize}
        \item The answer \answerNA{} means that the paper does not include experiments.
        \item If the paper includes experiments, a \answerNo{} answer to this question will not be perceived well by the reviewers: Making the paper reproducible is important, regardless of whether the code and data are provided or not.
        \item If the contribution is a dataset and\slash or model, the authors should describe the steps taken to make their results reproducible or verifiable. 
        \item Depending on the contribution, reproducibility can be accomplished in various ways. For example, if the contribution is a novel architecture, describing the architecture fully might suffice, or if the contribution is a specific model and empirical evaluation, it may be necessary to either make it possible for others to replicate the model with the same dataset, or provide access to the model. In general. releasing code and data is often one good way to accomplish this, but reproducibility can also be provided via detailed instructions for how to replicate the results, access to a hosted model (e.g., in the case of a large language model), releasing of a model checkpoint, or other means that are appropriate to the research performed.
        \item While NeurIPS does not require releasing code, the conference does require all submissions to provide some reasonable avenue for reproducibility, which may depend on the nature of the contribution. For example
        \begin{enumerate}
            \item If the contribution is primarily a new algorithm, the paper should make it clear how to reproduce that algorithm.
            \item If the contribution is primarily a new model architecture, the paper should describe the architecture clearly and fully.
            \item If the contribution is a new model (e.g., a large language model), then there should either be a way to access this model for reproducing the results or a way to reproduce the model (e.g., with an open-source dataset or instructions for how to construct the dataset).
            \item We recognize that reproducibility may be tricky in some cases, in which case authors are welcome to describe the particular way they provide for reproducibility. In the case of closed-source models, it may be that access to the model is limited in some way (e.g., to registered users), but it should be possible for other researchers to have some path to reproducing or verifying the results.
        \end{enumerate}
    \end{itemize}

\item {\bf Open access to data and code}
    \item[] Question: Does the paper provide open access to the data and code, with sufficient instructions to faithfully reproduce the main experimental results, as described in supplemental material?
    \item[] Answer: \answerYes{} 
    \item[] Justification: Code and data splits are available at the URL in the abstract;  datasets requiring data use agreements are linked to their original sources.
    \item[] Guidelines:
    \begin{itemize}
        \item The answer \answerNA{} means that paper does not include experiments requiring code.
        \item Please see the NeurIPS code and data submission guidelines (\url{https://neurips.cc/public/guides/CodeSubmissionPolicy}) for more details.
        \item While we encourage the release of code and data, we understand that this might not be possible, so \answerNo{} is an acceptable answer. Papers cannot be rejected simply for not including code, unless this is central to the contribution (e.g., for a new open-source benchmark).
        \item The instructions should contain the exact command and environment needed to run to reproduce the results. See the NeurIPS code and data submission guidelines (\url{https://neurips.cc/public/guides/CodeSubmissionPolicy}) for more details.
        \item The authors should provide instructions on data access and preparation, including how to access the raw data, preprocessed data, intermediate data, and generated data, etc.
        \item The authors should provide scripts to reproduce all experimental results for the new proposed method and baselines. If only a subset of experiments are reproducible, they should state which ones are omitted from the script and why.
        \item At submission time, to preserve anonymity, the authors should release anonymized versions (if applicable).
        \item Providing as much information as possible in supplemental material (appended to the paper) is recommended, but including URLs to data and code is permitted.
    \end{itemize}

\item {\bf Experimental setting/details}
    \item[] Question: Does the paper specify all the training and test details (e.g., data splits, hyperparameters, how they were chosen, type of optimizer) necessary to understand the results?
    \item[] Answer: \answerYes{} 
    \item[] Justification: All relevant training and testing details, including data splits, hyperparameter settings and selection process, optimizers used, and other experimental protocols, are provided in Section 4 and the supplementary material.
    \item[] Guidelines:
    \begin{itemize}
        \item The answer \answerNA{} means that the paper does not include experiments.
        \item The experimental setting should be presented in the core of the paper to a level of detail that is necessary to appreciate the results and make sense of them.
        \item The full details can be provided either with the code, in appendix, or as supplemental material.
    \end{itemize}

\item {\bf Experiment statistical significance}
    \item[] Question: Does the paper report error bars suitably and correctly defined or other appropriate information about the statistical significance of the experiments?
    \item[] Answer: \answerYes{} 
    \item[] Justification: We report error bars and/or standard deviations for our experimental results, following standard statistical practices, as described in Section 4.
    \item[] Guidelines:
    \begin{itemize}
        \item The answer \answerNA{} means that the paper does not include experiments.
        \item The authors should answer \answerYes{} if the results are accompanied by error bars, confidence intervals, or statistical significance tests, at least for the experiments that support the main claims of the paper.
        \item The factors of variability that the error bars are capturing should be clearly stated (for example, train/test split, initialization, random drawing of some parameter, or overall run with given experimental conditions).
        \item The method for calculating the error bars should be explained (closed form formula, call to a library function, bootstrap, etc.)
        \item The assumptions made should be given (e.g., Normally distributed errors).
        \item It should be clear whether the error bar is the standard deviation or the standard error of the mean.
        \item It is OK to report 1-sigma error bars, but one should state it. The authors should preferably report a 2-sigma error bar than state that they have a 96\% CI, if the hypothesis of Normality of errors is not verified.
        \item For asymmetric distributions, the authors should be careful not to show in tables or figures symmetric error bars that would yield results that are out of range (e.g., negative error rates).
        \item If error bars are reported in tables or plots, the authors should explain in the text how they were calculated and reference the corresponding figures or tables in the text.
    \end{itemize}

\item {\bf Experiments compute resources}
    \item[] Question: For each experiment, does the paper provide sufficient information on the computer resources (type of compute workers, memory, time of execution) needed to reproduce the experiments?
    \item[] Answer: \answerYes{} 
    \item[] Justification: We provide detailed information on the compute environment used for our experiments, including hardware specifications (e.g., GPU type, RAM), and approximate execution time, in the supplementary material.
    \item[] Guidelines:
    \begin{itemize}
        \item The answer \answerNA{} means that the paper does not include experiments.
        \item The paper should indicate the type of compute workers CPU or GPU, internal cluster, or cloud provider, including relevant memory and storage.
        \item The paper should provide the amount of compute required for each of the individual experimental runs as well as estimate the total compute. 
        \item The paper should disclose whether the full research project required more compute than the experiments reported in the paper (e.g., preliminary or failed experiments that didn't make it into the paper). 
    \end{itemize}
    
\item {\bf Code of ethics}
    \item[] Question: Does the research conducted in the paper conform, in every respect, with the NeurIPS Code of Ethics \url{https://neurips.cc/public/EthicsGuidelines}?
    \item[] Answer: \answerYes{} 
    \item[] Justification: Our research fully adheres to the NeurIPS Code of Ethics, as we have ensured that all experiments and data usage comply with ethical guidelines and respect privacy, fairness, and transparency throughout our work.
    \item[] Guidelines:
    \begin{itemize}
        \item The answer \answerNA{} means that the authors have not reviewed the NeurIPS Code of Ethics.
        \item If the authors answer \answerNo, they should explain the special circumstances that require a deviation from the Code of Ethics.
        \item The authors should make sure to preserve anonymity (e.g., if there is a special consideration due to laws or regulations in their jurisdiction).
    \end{itemize}

\item {\bf Broader impacts}
    \item[] Question: Does the paper discuss both potential positive societal impacts and negative societal impacts of the work performed?
    \item[] Answer: \answerYes{} 
    \item[] Justification: We discuss the broader societal impacts of our work in Section 5, including how improved privacy evaluation can guide safer deployment of DP synthetic data methods (positive), and the risk that uncovering attack techniques may also inform potential adversaries (negative).
    \item[] Guidelines:
    \begin{itemize}
        \item The answer \answerNA{} means that there is no societal impact of the work performed.
        \item If the authors answer \answerNA{} or \answerNo, they should explain why their work has no societal impact or why the paper does not address societal impact.
        \item Examples of negative societal impacts include potential malicious or unintended uses (e.g., disinformation, generating fake profiles, surveillance), fairness considerations (e.g., deployment of technologies that could make decisions that unfairly impact specific groups), privacy considerations, and security considerations.
        \item The conference expects that many papers will be foundational research and not tied to particular applications, let alone deployments. However, if there is a direct path to any negative applications, the authors should point it out. For example, it is legitimate to point out that an improvement in the quality of generative models could be used to generate Deepfakes for disinformation. On the other hand, it is not needed to point out that a generic algorithm for optimizing neural networks could enable people to train models that generate Deepfakes faster.
        \item The authors should consider possible harms that could arise when the technology is being used as intended and functioning correctly, harms that could arise when the technology is being used as intended but gives incorrect results, and harms following from (intentional or unintentional) misuse of the technology.
        \item If there are negative societal impacts, the authors could also discuss possible mitigation strategies (e.g., gated release of models, providing defenses in addition to attacks, mechanisms for monitoring misuse, mechanisms to monitor how a system learns from feedback over time, improving the efficiency and accessibility of ML).
    \end{itemize}
    
\item {\bf Safeguards}
    \item[] Question: Does the paper describe safeguards that have been put in place for responsible release of data or models that have a high risk for misuse (e.g., pre-trained language models, image generators, or scraped datasets)?
    \item[] Answer: \answerNA{} 
    \item[] Justification: This paper does not release any new models or datasets with a high risk for misuse; all data and models used are from existing publicly available sources.
    \item[] Guidelines:
    \begin{itemize}
        \item The answer \answerNA{} means that the paper poses no such risks.
        \item Released models that have a high risk for misuse or dual-use should be released with necessary safeguards to allow for controlled use of the model, for example by requiring that users adhere to usage guidelines or restrictions to access the model or implementing safety filters. 
        \item Datasets that have been scraped from the Internet could pose safety risks. The authors should describe how they avoided releasing unsafe images.
        \item We recognize that providing effective safeguards is challenging, and many papers do not require this, but we encourage authors to take this into account and make a best faith effort.
    \end{itemize}

\item {\bf Licenses for existing assets}
    \item[] Question: Are the creators or original owners of assets (e.g., code, data, models), used in the paper, properly credited and are the license and terms of use explicitly mentioned and properly respected?
    \item[] Answer: \answerYes{} 
    \item[] Justification: All external datasets, code, and models used in this work are properly credited in the paper, and their licenses and terms of use are explicitly acknowledged and respected, as detailed in the text and the references.
    \item[] Guidelines:
    \begin{itemize}
        \item The answer \answerYes{} means that the paper does not use existing assets.
        \item The authors should cite the original paper that produced the code package or dataset.
        \item The authors should state which version of the asset is used and, if possible, include a URL.
        \item The name of the license (e.g., CC-BY 4.0) should be included for each asset.
        \item For scraped data from a particular source (e.g., website), the copyright and terms of service of that source should be provided.
        \item If assets are released, the license, copyright information, and terms of use in the package should be provided. For popular datasets, \url{paperswithcode.com/datasets} has curated licenses for some datasets. Their licensing guide can help determine the license of a dataset.
        \item For existing datasets that are re-packaged, both the original license and the license of the derived asset (if it has changed) should be provided.
        \item If this information is not available online, the authors are encouraged to reach out to the asset's creators.
    \end{itemize}

\item {\bf New assets}
    \item[] Question: Are new assets introduced in the paper well documented and is the documentation provided alongside the assets?
    \item[] Answer: \answerYes{} 
    \item[] Justification: We introduce a benchmark and an auditing setup for synthetic text, and provide the corresponding code with clear documentation and usage instructions in the supplementary material and online repository.
    \item[] Guidelines:
    \begin{itemize}
        \item The answer \answerNA{} means that the paper does not release new assets.
        \item Researchers should communicate the details of the dataset\slash code\slash model as part of their submissions via structured templates. This includes details about training, license, limitations, etc. 
        \item The paper should discuss whether and how consent was obtained from people whose asset is used.
        \item At submission time, remember to anonymize your assets (if applicable). You can either create an anonymized URL or include an anonymized zip file.
    \end{itemize}

\item {\bf Crowdsourcing and research with human subjects}
    \item[] Question: For crowdsourcing experiments and research with human subjects, does the paper include the full text of instructions given to participants and screenshots, if applicable, as well as details about compensation (if any)? 
    \item[] Answer: \answerNA{} 
    \item[] Justification: This work does not involve any crowdsourcing experiments or research with human subjects.
    \item[] Guidelines:
    \begin{itemize}
        \item The answer \answerNA{} means that the paper does not involve crowdsourcing nor research with human subjects.
        \item Including this information in the supplemental material is fine, but if the main contribution of the paper involves human subjects, then as much detail as possible should be included in the main paper. 
        \item According to the NeurIPS Code of Ethics, workers involved in data collection, curation, or other labor should be paid at least the minimum wage in the country of the data collector. 
    \end{itemize}

\item {\bf Institutional review board (IRB) approvals or equivalent for research with human subjects}
    \item[] Question: Does the paper describe potential risks incurred by study participants, whether such risks were disclosed to the subjects, and whether Institutional Review Board (IRB) approvals (or an equivalent approval/review based on the requirements of your country or institution) were obtained?
    \item[] Answer: \answerNA{} 
    \item[] Justification: This work does not involve research with human subjects and therefore IRB approval was not required.
    \item[] Guidelines:
    \begin{itemize}
        \item The answer \answerNA{} means that the paper does not involve crowdsourcing nor research with human subjects.
        \item Depending on the country in which research is conducted, IRB approval (or equivalent) may be required for any human subjects research. If you obtained IRB approval, you should clearly state this in the paper. 
        \item We recognize that the procedures for this may vary significantly between institutions and locations, and we expect authors to adhere to the NeurIPS Code of Ethics and the guidelines for their institution. 
        \item For initial submissions, do not include any information that would break anonymity (if applicable), such as the institution conducting the review.
    \end{itemize}

\item {\bf Declaration of LLM usage}
    \item[] Question: Does the paper describe the usage of LLMs if it is an important, original, or non-standard component of the core methods in this research? Note that if the LLM is used only for writing, editing, or formatting purposes and does \emph{not} impact the core methodology, scientific rigor, or originality of the research, declaration is not required.
    \item[] Answer: \answerYes{} 
    \item[] Justification: Our study evaluates the privacy protection of differentially private synthetic data generation methods using large language models (LLMs); the usage and configuration of LLMs in our experiments are detailed in Section 5 and the supplementary material.
    \item[] Guidelines:
    \begin{itemize}
        \item The answer \answerNA{} means that the core method development in this research does not involve LLMs as any important, original, or non-standard components.
        \item Please refer to our LLM policy in the NeurIPS handbook for what should or should not be described.
    \end{itemize}

\end{enumerate}

\clearpage

\appendix

\section{Supplemental Material}
\subsection{On DP-Inference}
\label{app:dp-inference}
We further investigate the feasibility of DP-inference methods for large-scale synthesis of domain-specific data. We select a representative algorithm~\cite{amin2024private} which alleviates the issues of previous DP-inference approaches~\cite{Tang2023Privacy-PreservingGeneration}, purportedly making it suitable to infer large datasets with DP guarantees. This is done by choosing to occasionally sample from a public LLM based on whether public and private (as in, containing private examples in their prompt) LLMs difference in predictions do not surpass a noisy threshold, setting the threshold using the sparse vector technique, instead of privatising the whole predicted distribution. We re-implement their algorithm and (disregarding label conditioning) attempt at generating a synthetic version of the \textsc{PsyTAR} dataset. We find that even at the lowest temperature setting of $\tau = 1.5$ no combination of noise/threshold settings reported in the paper yield coherent text. We lower $\tau$ to 0.5 until we obtain coherent text, and set $(\sigma, \theta)$ to $(0.5, 0.7)$. However, even at such large $\sigma$ values, the observed probability of sampling from the private model is around $90\%$, which means that we run out of tokens to generate due to privacy budget (as the privacy loss scales with each generated token), before we generate enough tokens to represent the whole dataset. We can reduce the privacy loss by increasing either \emph{(a)} the batch size, however, this is constrained by both practical GPU memory constraints as well as the size of the dataset (5000 points); or, \emph{(b)}, the number of batches, which is similarly constrained by the dataset size. Based on our observations, we conclude that DP inference is only suitable to generate a significantly smaller amount of data points than the original dataset---indeed~\cite{Tang2023Privacy-PreservingGeneration} themselves report synthetic dataset sizes of magnitudes smaller than the original dataset sizes.

\subsection{Additional Results}
\label{app:additional_results}

\subsubsection{Best-Classifier Results across Generator Sizes}
\label{app:size-best}
\begin{table}[h]
\centering
\footnotesize
\setlength{\tabcolsep}{3pt}
\begin{tabular}{llccccc}
\toprule
 & Size & $\varepsilon=\infty$ & $\varepsilon=4$ & $\varepsilon=2$ & $\varepsilon=1$ & $\varepsilon=0.5$ \\
\midrule
\multirow{3}{*}{\textsc{PsyTAR}}
 & 1B & 76.2\pmci{0.2} & 35.4\pmci{3.6} & 36.9\pmci{3.2} & 35.2\pmci{4.4} & 32.5\pmci{2.2} \\
 & 3B & 76.4\pmci{1.0} & 41.1\pmci{1.4} & 37.1\pmci{2.5} & 31.9\pmci{3.9} & 25.2\pmci{3.9} \\
 & 8B & \textbf{77.4}\pmci{0.6} & \textbf{43.1}\pmci{2.9} & \textbf{43.9}\pmci{2.2} & \textbf{42.0}\pmci{5.4} & \textbf{35.6}\pmci{1.9} \\
\midrule
\multirow{3}{*}{\textsc{N2C2'08}}
 & 1B & 59.1\pmci{14.0} & 68.6\pmci{1.0} & 69.9\pmci{3.4} & \textbf{71.7}\pmci{3.6} & \textbf{72.7}\pmci{1.8} \\
 & 3B & 65.9\pmci{15.0} & \textbf{70.8}\pmci{7.8} & \textbf{73.0}\pmci{2.2} & 70.3\pmci{3.3} & 70.6\pmci{3.3} \\
 & 8B & \textbf{76.4}\pmci{8.4} & 69.6\pmci{1.5} & 70.7\pmci{2.2} & 71.5\pmci{6.2} & 69.5\pmci{5.6} \\
\midrule
\multirow{3}{*}{\textsc{DMSAFN}}
 & 1B & 82.6\pmci{0.4} & 45.3\pmci{10.2} & 46.0\pmci{10.4} & 43.1\pmci{5.8} & 55.7\pmci{10.8} \\
 & 3B & \textbf{86.8}\pmci{6.2} & \textbf{46.5}\pmci{11.3} & \textbf{47.9}\pmci{11.9} & 43.5\pmci{4.6} & \textbf{73.4}\pmci{11.9} \\
 & 8B & 86.4\pmci{1.3} & 28.3\pmci{22.2} & 39.9\pmci{8.9} & \textbf{45.5}\pmci{8.7} & 53.3\pmci{6.5} \\
\midrule
\multirow{3}{*}{\textsc{AsyLax}}
 & 1B & 36.2\pmci{19.8} & \textbf{50.9}\pmci{1.8} & 47.8\pmci{4.0} & \textbf{45.3}\pmci{4.4} & \textbf{40.8}\pmci{0.4} \\
 & 3B & 49.1\pmci{6.7} & 42.9\pmci{4.6} & \textbf{49.3}\pmci{2.2} & 44.7\pmci{9.1} & 39.1\pmci{22.7} \\
 & 8B & \textbf{51.4}\pmci{0.0} & 30.8\pmci{7.6} & 28.4\pmci{3.0} & 27.6\pmci{0.1} & 27.7\pmci{0.1} \\
\bottomrule
\end{tabular}
\caption{Micro-F1 (best classifier) of synthetic data from DP-fine-tuned Llama models of increasing size. Subscripts: 95\% confidence interval over 3 seeds.}
\label{tab:text-llama-utility-best}
\end{table}

Table~\ref{tab:text-llama-utility-best} reports the best-classifier
counterpart of Table~\ref{tab:text-llama-utility}. The qualitative picture
is unchanged, indicating that our conclusions on generator size are robust
to the choice of downstream classifier. Without privacy constraints, larger
generators generally yield higher utility (e.g., \textsc{N2C2'08}:
$59.1 \rightarrow 76.4$ from 1B to 8B), although on \textsc{DMSAFN} the 3B
and 8B models perform on par. Under DP, no consistent size ordering
emerges: the 8B model is best at every $\varepsilon$ on \textsc{PsyTAR},
the 3B model is best in most settings on \textsc{DMSAFN}, and the smaller
models are best on \textsc{AsyLax}, while on \textsc{N2C2'08} the
differences between sizes are mostly within the confidence intervals.

Two results warrant caution. On \textsc{AsyLax}, the 8B model at
$\varepsilon=\infty$ attains $51.4$ with zero variance, which equals the
majority-class baseline, and at $\varepsilon \leq 1$ it drops to
$27.6$--$27.7$ with near-zero variance; both patterns suggest degenerate
synthetic data rather than genuine differences in utility. In addition,
several settings show wide confidence intervals (e.g., up to $\pm 22.2$ on
\textsc{DMSAFN} and $\pm 22.7$ on \textsc{AsyLax}), so differences between
sizes should be interpreted with care given only three seeds.

{
\subsubsection{Results with OLMo-2}
\label{app:olmo}

\paragraph{Motivation.}
To test whether our findings extend beyond the Llama family, we repeat the evaluation with a second generator backbone, \textsc{OLMo-2-0425-1B}.
Because its pre-training corpora are publicly released, OLMo-2 also lets us measure contamination exactly rather than through the RedPajama proxy used for Llama, which we exploit  in a controlled experiment relating pre-training leakage to downstream utility.

\paragraph{Setup.}
\textbf{For the utility results} (Table~\ref{tab:utility-olmo}), we follow the same protocol as the main Llama experiments, replacing only the generator backbone with \textsc{OLMo-2-0425-1B}; all
other settings, including datasets, privacy budgets, downstream classifiers, and seeds, are unchanged.
\textbf{For the contamination analysis}, we use the fact that \textsc{OLMo-2-0425-1B} is pre-trained on \textsc{olmo-mix-1124} (stage~1) and mid-trained on \textsc{Dolmino-mix-1124} (stage~2), both
publicly released.
We check every HoC training sentence against both corpora. For \textsc{olmo-mix-1124}, a sentence is flagged if its full text occurs verbatim in the corpus (Infini-gram exact match, after detokenizing the pre-tokenized HoC text). 

DOLMINO-MIX-1124 has no Infini-gram index, so we scan all 732M of its documents for the same full-sentence match. We compare text after lower-casing and normalising whitespace. A match counts as specific to HoC only if the matching document also contains at least one other complete HoC training sentence, i.e. the document reproduces the source abstract (in practice mostly PES2O papers). Short sentences whose text also occurs in a sample of generic web text (DCLM) count as generic phrases, and sentences too short to match reliably count as unverifiable.

In total, 58\% of the HoC training sentences are flagged by at least one corpus. We use HoC because this rate is close to one half, which leaves enough sentences on both sides to vary the leaked share of a training set of fixed size. 
We split the training set into a \emph{leaked} pool of 5{,}360 sentences that occur in \textsc{Dolmino-mix-1124} within documents that copy their source abstracts, and a \emph{clean} pool of 4{,}299 sentences found in neither corpus; the remaining 642 sentences (e.g.\ found
only in \textsc{olmo-mix-1124}, or matched only as generic phrases) are not used. Each private training set has $N = 4{,}000$ sentences, of which a fraction $x$ is drawn from the leaked pool and $1-x$ from the clean pool, for $x \in \{0, 0.01, \dots, 1\}$.
For each training set we fully fine-tune OLMo-2-0425-1B with DP-SGD. We use per-example clipping norm 1.0, batch size 1,024, 10 epochs, sequence length 128 and $\delta \frac{1}{N^2} = 6.25 \times 10^{-8}$. At $\epsilon = 4$ we train three independent replicates, each with its own random draw of the 101 training sets (303 generators). Each generator produces one synthetic sentence per training sentence, prompted with that sentence's labels.  For utility, we train BiomedBERT-large (its HoC configuration) and BioLinkBERT-large (the same configuration) on every synthetic set with three seeds, and scored by best-epoch micro-F1 on the real HoC test set. Runs that predict no hallmark for any test sentence (micro-F1 = 0) are left out (1 of 909 BiomedBERT runs and 162 of 909 BioLinkBERT runs at $\epsilon=4$). We fit an ordinary least-squares slope of utility on leakage. We test slope > 0 with a one-sided permutation test: the points are shuffled within each replicate. The overall $p$ tests the four views jointly: it compares the weakest of the four standardised slopes with its distribution under the same shuffles. Figure~\ref{fig:olmo-hoc-contamination} plots utility against x for mean and max scores across the three training runs per leakage percentage, for both downstream models, respectively.

\paragraph{Utility across privacy budgets.}
\begin{table*}[h]
\centering
\small
\setlength{\tabcolsep}{4pt}
\begin{tabular}{llcccccc}
\toprule
Dataset & Classifier & Original & $\epsilon=\infty$ & $\epsilon=0.5$ & $\epsilon=1$ & $\epsilon=2$ & $\epsilon=4$ \\
\midrule
\multirow{3}{*}{\shortstack[l]{HoC\\{\scriptsize R 3.7 / M 9.1}}} & BERT-large & 71.7{\scriptsize$\pm$2.2} & 65.3{\scriptsize$\pm$1.8} & 27.6{\scriptsize$\pm$6.4} & 25.0{\scriptsize$\pm$2.9} & 25.2{\scriptsize$\pm$4.1} & 25.5{\scriptsize$\pm$6.1} \\
 & BioClinicalBERT & 67.6{\scriptsize$\pm$5.8} & 61.4{\scriptsize$\pm$3.3} & 25.9{\scriptsize$\pm$2.5} & 23.8{\scriptsize$\pm$2.8} & 25.9{\scriptsize$\pm$2.2} & 22.0{\scriptsize$\pm$2.9} \\
 & BiomedBERT & 74.0{\scriptsize$\pm$0.7} & 65.4{\scriptsize$\pm$1.3} & 29.4{\scriptsize$\pm$4.0} & 26.4{\scriptsize$\pm$2.0} & 29.0{\scriptsize$\pm$1.0} & 28.6{\scriptsize$\pm$1.9} \\
\midrule
\multirow{3}{*}{\shortstack[l]{N2C2'08\\{\scriptsize R 46.1 / M 53.2}}} & LongFormer-large & 82.6{\scriptsize$\pm$14.5} & 68.9{\scriptsize$\pm$22.0} & 55.9{\scriptsize$\pm$8.4} & 54.0{\scriptsize$\pm$6.5} & 55.1{\scriptsize$\pm$5.5} & 56.0{\scriptsize$\pm$12.6} \\
 & Clinical-BigBird & 87.4{\scriptsize$\pm$2.4} & 72.7{\scriptsize$\pm$6.2} & 52.0{\scriptsize$\pm$12.4} & 59.6{\scriptsize$\pm$1.9} & 52.0{\scriptsize$\pm$10.9} & 50.3{\scriptsize$\pm$10.5} \\
 & Clinical-LongFormer & 77.9{\scriptsize$\pm$43.5} & 73.6{\scriptsize$\pm$9.0} & 52.4{\scriptsize$\pm$1.0} & 54.9{\scriptsize$\pm$2.1} & 53.1{\scriptsize$\pm$11.9} & 54.4{\scriptsize$\pm$4.5} \\
\midrule
\multirow{3}{*}{\shortstack[l]{PsyTAR\\{\scriptsize R 25.4 / M 41.8}}} & BERT-base & 78.6{\scriptsize$\pm$0.6} & 76.6{\scriptsize$\pm$1.2} & 17.9{\scriptsize$\pm$3.6} & 19.2{\scriptsize$\pm$0.7} & 17.7{\scriptsize$\pm$3.0} & 21.9{\scriptsize$\pm$3.8} \\
 & BERT-large & 79.5{\scriptsize$\pm$1.6} & 77.7{\scriptsize$\pm$3.1} & 16.7{\scriptsize$\pm$4.0} & 18.6{\scriptsize$\pm$2.2} & 18.4{\scriptsize$\pm$6.2} & 21.8{\scriptsize$\pm$5.6} \\
 & DeBERTa-xlarge & 82.1{\scriptsize$\pm$1.1} & 80.1{\scriptsize$\pm$0.9} & 11.0{\scriptsize$\pm$7.1} & 22.2{\scriptsize$\pm$1.2} & 13.4{\scriptsize$\pm$11.7} & 21.6{\scriptsize$\pm$2.1} \\
\midrule
\multirow{3}{*}{\shortstack[l]{Mimic\\{\scriptsize R 48.8 / M 52.5}}} & BiomedBERT & 76.3{\scriptsize$\pm$1.2} & 72.4{\scriptsize$\pm$1.7} & 55.1{\scriptsize$\pm$9.7} & 56.9{\scriptsize$\pm$5.9} & 59.7{\scriptsize$\pm$2.1} & 55.4{\scriptsize$\pm$14.6} \\
 & BioClinicalBERT & 74.8{\scriptsize$\pm$0.2} & 71.6{\scriptsize$\pm$0.8} & 59.1{\scriptsize$\pm$1.2} & 56.3{\scriptsize$\pm$6.8} & 64.2{\scriptsize$\pm$3.4} & 57.2{\scriptsize$\pm$15.0} \\
 & BiomedRobertaBase & 73.4{\scriptsize$\pm$0.9} & 71.3{\scriptsize$\pm$0.4} & 39.9{\scriptsize$\pm$11.6} & 35.5{\scriptsize$\pm$25.9} & 47.7{\scriptsize$\pm$9.6} & 37.9{\scriptsize$\pm$10.6} \\
\midrule
\multirow{3}{*}{\shortstack[l]{DMSAFN\\{\scriptsize R 30.5 / M 41.1}}} & ProsusAI-finbert & 79.6{\scriptsize$\pm$6.4} & 77.8{\scriptsize$\pm$7.4} & 46.5{\scriptsize$\pm$24.7} & 47.6{\scriptsize$\pm$26.8} & 39.6{\scriptsize$\pm$16.9} & 37.7{\scriptsize$\pm$3.1} \\
 & finbert-tone & 73.4{\scriptsize$\pm$2.8} & 73.3{\scriptsize$\pm$4.7} & 34.4{\scriptsize$\pm$7.1} & 38.9{\scriptsize$\pm$16.9} & 36.4{\scriptsize$\pm$7.7} & 32.4{\scriptsize$\pm$10.1} \\
 & FinancialBERT & 92.0{\scriptsize$\pm$3.9} & 89.7{\scriptsize$\pm$8.2} & 51.0{\scriptsize$\pm$11.7} & 57.2{\scriptsize$\pm$11.8} & 41.2{\scriptsize$\pm$15.8} & 35.4{\scriptsize$\pm$6.3} \\
\midrule
\multirow{3}{*}{\shortstack[l]{Lc37Fnd\\{\scriptsize R 41.3 / M 50.2}}} & ProsusAI-finbert & 86.7{\scriptsize$\pm$0.1} & 68.4{\scriptsize$\pm$0.4} & 54.1{\scriptsize$\pm$7.9} & 54.5{\scriptsize$\pm$7.6} & 55.1{\scriptsize$\pm$3.9} & 54.2{\scriptsize$\pm$2.5} \\
 & distilroberta & 91.1{\scriptsize$\pm$1.8} & 72.9{\scriptsize$\pm$2.6} & 57.7{\scriptsize$\pm$4.7} & 58.1{\scriptsize$\pm$0.3} & 59.2{\scriptsize$\pm$1.1} & 57.2{\scriptsize$\pm$3.2} \\
 & FinancialBERT & 82.7{\scriptsize$\pm$1.9} & 68.0{\scriptsize$\pm$1.8} & 56.9{\scriptsize$\pm$4.3} & 57.8{\scriptsize$\pm$5.3} & 55.8{\scriptsize$\pm$2.8} & 49.3{\scriptsize$\pm$14.9} \\
\midrule
\multirow{3}{*}{\shortstack[l]{ZsTfns\\{\scriptsize R 48.7 / M 65.6}}} & ProsusAI-finbert & 88.9{\scriptsize$\pm$0.7} & 86.7{\scriptsize$\pm$1.9} & 47.6{\scriptsize$\pm$5.2} & 52.6{\scriptsize$\pm$6.4} & 57.5{\scriptsize$\pm$12.3} & 59.2{\scriptsize$\pm$1.9} \\
 & finbert-tone & 87.6{\scriptsize$\pm$1.3} & 85.9{\scriptsize$\pm$1.5} & 47.1{\scriptsize$\pm$1.0} & 52.6{\scriptsize$\pm$5.8} & 54.8{\scriptsize$\pm$9.6} & 57.5{\scriptsize$\pm$3.0} \\
 & FinancialBERT & 84.3{\scriptsize$\pm$1.1} & 81.6{\scriptsize$\pm$1.1} & 50.6{\scriptsize$\pm$10.2} & 56.6{\scriptsize$\pm$7.3} & 60.1{\scriptsize$\pm$8.6} & 61.8{\scriptsize$\pm$7.7} \\
\midrule
\multirow{3}{*}{\shortstack[l]{Asylax\\{\scriptsize R 38.4 / M 51.4}}} & legal-bert-base & 65.6{\scriptsize$\pm$5.0} & 58.4{\scriptsize$\pm$3.5} & 44.2{\scriptsize$\pm$9.0} & 34.6{\scriptsize$\pm$24.8} & 38.4{\scriptsize$\pm$11.9} & 37.5{\scriptsize$\pm$26.4} \\
 & InLegalBERT & 65.3{\scriptsize$\pm$0.7} & 58.0{\scriptsize$\pm$4.5} & 36.9{\scriptsize$\pm$5.5} & 34.6{\scriptsize$\pm$26.1} & 30.4{\scriptsize$\pm$5.6} & 33.8{\scriptsize$\pm$11.0} \\
 & Roberta-base & 58.5{\scriptsize$\pm$2.1} & 58.2{\scriptsize$\pm$3.7} & 35.2{\scriptsize$\pm$15.7} & 30.7{\scriptsize$\pm$7.2} & 29.5{\scriptsize$\pm$2.0} & 28.8{\scriptsize$\pm$0.7} \\
\midrule
\multirow{3}{*}{\shortstack[l]{Eurlex\\{\scriptsize R 24.1 / M 36.1}}} & legal-bert-base & 89.9{\scriptsize$\pm$0.6} & 86.3{\scriptsize$\pm$0.9} & 33.6{\scriptsize$\pm$5.2} & 36.6{\scriptsize$\pm$6.4} & 38.5{\scriptsize$\pm$1.9} & 38.9{\scriptsize$\pm$8.3} \\
 & InLegalBERT & 89.8{\scriptsize$\pm$0.6} & 86.3{\scriptsize$\pm$0.8} & 33.8{\scriptsize$\pm$5.0} & 39.6{\scriptsize$\pm$2.9} & 41.7{\scriptsize$\pm$10.9} & 44.1{\scriptsize$\pm$7.0} \\
 & Roberta-base & 89.3{\scriptsize$\pm$0.8} & 86.0{\scriptsize$\pm$0.3} & 39.9{\scriptsize$\pm$4.0} & 39.1{\scriptsize$\pm$4.3} & 38.8{\scriptsize$\pm$11.3} & 41.9{\scriptsize$\pm$12.6} \\
\bottomrule
\end{tabular}
\caption{Downstream classification micro-F1 scores with synthetic data from OLMo-2-0425-1B (fine-tuned for 10 epochs) generated with DP-Gen, across privacy budgets ($\epsilon$); ``Original'' trains on the real data. R / M: random / majority guess baselines. Each value is the mean over 3 seeds $\pm$ the 95\% confidence interval (t-distribution).}
\label{tab:utility-olmo}
\end{table*}

Table~\ref{tab:utility-olmo} shows that, as with Llama, utility drops sharply once DP is applied, and on several datasets (\textsc{PsyTAR}, \textsc{AsyLax}, \textsc{ZsTfns}, \textsc{DMSAFN}) DP-Gen falls to or below the stronger of the random and majority baselines. Across datasets, the retained utility does not track estimated contamination: the most contaminated datasets (\textsc{DMSAFN} and \textsc{Eurlex}) barely exceed the majority baseline at $\varepsilon=2$, whereas \textsc{HoC} retains the most. This is consistent with contamination being only one of several dataset-level factors, alongside dataset size, label space, and task difficulty, and motivates the controlled experiment below, which varies contamination within a single dataset.

\begin{figure}[ht]
  \centering
  \includegraphics[width=\columnwidth]{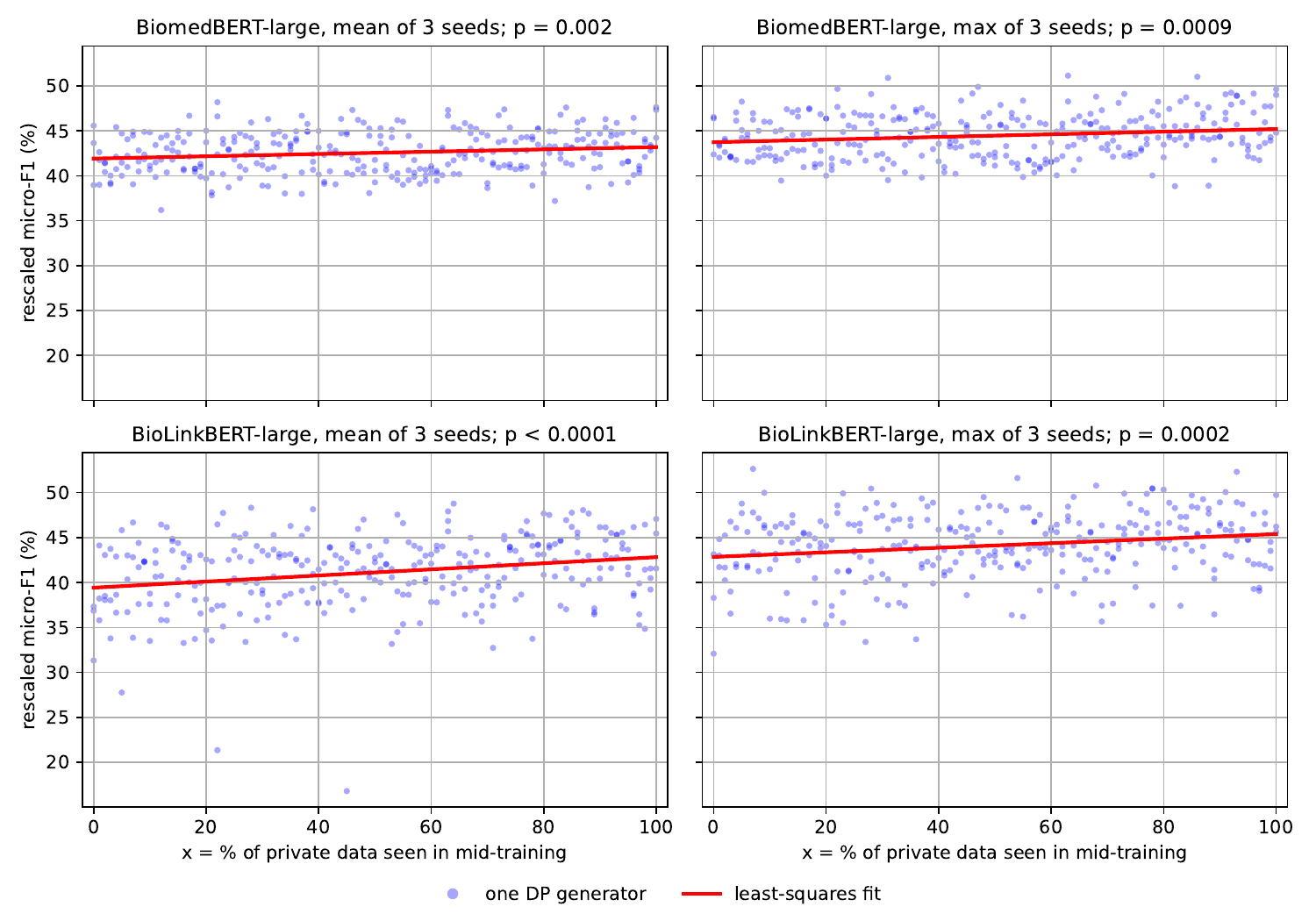}
  \caption{Utility of OLMo-2-1B DP synthetic data ($\varepsilon = 4$) for HoC against the share $x$ of the
  private training set ($N = 4{,}000$) that OLMo-2 saw in mid-training (dolmino-mix-1124). Each point is one DP
  generator (303 from three independent replicates), scored by a classifier trained on its synthetic data and
  tested on the real HoC test set, as the mean or maximum over 3 classifier seeds. Micro-F1 is rescaled per
  classifier to the percentage of the gain over random guessing (0.037) that the same classifier achieves when
  trained on the real 4{,}000 records (0.719 BiomedBERT, 0.712 BioLinkBERT). Red: least-squares fit; $p$:
  one-sided permutation test of a positive slope.}
  \label{fig:olmo-hoc-contamination}
\end{figure}

\begin{table}[t]
\centering
\footnotesize
\setlength{\tabcolsep}{4pt}
\begin{tabular}{lcccc}
\toprule
Classifier (seeds) & $x=0 \to 100$ & $\Delta$ & $p$ & $p$ (collapsed $=0$) \\
\midrule
BiomedBERT (mean) & 41.9 $\to$ 43.2 & +1.3 & 0.002 & 0.003 \\
BiomedBERT (max) & 43.7 $\to$ 45.2 & +1.5 & 0.0009 & 0.0009 \\
BioLinkBERT (mean) & 39.4 $\to$ 42.8 & +3.4 & $<$0.0001 & 0.009 \\
BioLinkBERT (max) & 42.9 $\to$ 45.4 & +2.5 & 0.0002 & 0.0003 \\
\midrule
\emph{all four (overall)} & & & $<$0.0001 & 0.0001 \\
\bottomrule
\end{tabular}
\caption{Rescaled downstream micro-F1 (\%) of OLMo-2-1B DP synthetic data ($\varepsilon = 4$, 303 generators) for HoC as the share $x$ of private records seen in OLMo-2 mid-training grows from 0\% to 100\% ($N = 4{,}000$): least-squares fit at the two ends, its change, and one-sided permutation $p$ of a positive slope (20{,}000 shuffles of $x$ within each of the three replicates). Collapsed classifier runs (micro-F1 0) are left out, or counted as 0 in the last column. Overall: one test of all four views at once (weakest standardised slope).}
\label{tab:olmo-hoc-contamination}
\end{table}

\paragraph{Pre-training exposure and utility.}
Figure~\ref{fig:olmo-hoc-contamination} and Table~\ref{tab:olmo-hoc-contamination} report the replacement experiment on HoC at $\varepsilon = 4$. 
To compare classifiers on one scale, we rescale micro-F1 to the share of the real-data gain over random guessing (0\% = random guessing at the training-set label frequencies; 100\% = the
same classifier trained on the real 4{,}000 records). 
Raising the share of private records that OLMo-2 saw in mid-training from 0\% to 100\% raises this share by 1.3--3.4 points, from about 39--44\% to 43--45\% (one-sided permutation $p \leq 0.002$ for each of the four views; overall $p < 0.0001$), with a similar slope in each of the three independent replicates. 
}

\subsubsection{Cross-attack MIA audit on SYN vs.\ PRI}
\label{app:mia_tables}

Table~\ref{tab:mia_attacks_syn_priv_app} reports AUC for three attacks (RMIA, LiRA, logits-based) under four regimes: downstream classifier trained on synthetic data (\textsc{SYN}) or on private data (\textsc{PRI}), each with/without DP (``No DP'' vs.\ $\epsilon=2$). Higher AUC indicates stronger membership distinguishability ($0.5$ is chance). These results complement the main-text discussion on attack sensitivity and the interaction between generator pre-training exposure and downstream privacy risk.

\begin{table}[h]
\centering
\small
\caption{\textbf{MIA AUC across attacks and training regimes.} \textsc{SYN} denotes a downstream classifier trained on synthetic data; \textsc{PRI} denotes training on private data. ``No DP'' vs.\ $\epsilon=2$ refers to whether DP is applied in the relevant training/generation step described in Section~\ref{sec:experiments}. Higher AUC indicates stronger membership distinguishability ($0.5$ is chance).}
\label{tab:mia_attacks_syn_priv_app}
\begin{tabular}{llccc}
\toprule
\textbf{Dataset} & \textbf{Regime} & \textbf{RMIA} & \textbf{LiRA} & \textbf{Logits} \\
\midrule
\textsc{DMSAFN} & \textsc{SYN} (No DP) & 0.60 & 0.49 & 0.60 \\
\textsc{DMSAFN} & \textsc{SYN} ($\epsilon=2$) & 0.54 & 0.45 & 0.48 \\
\textsc{DMSAFN} & \textsc{PRI} (No DP) & 0.45 & 0.50 & 0.54 \\
\textsc{DMSAFN} & \textsc{PRI} ($\epsilon=2$) & 0.49 & 0.49 & 0.54 \\
\midrule
\textsc{PsyTAR} & \textsc{SYN} (No DP) & 0.65 & 0.61 & 0.62 \\
\textsc{PsyTAR} & \textsc{SYN} ($\epsilon=2$) & 0.57 & 0.61 & 0.61 \\
\textsc{PsyTAR} & \textsc{PRI} (No DP) & 0.45 & 0.48 & 0.45 \\
\textsc{PsyTAR} & \textsc{PRI} ($\epsilon=2$) & 0.52 & 0.57 & 0.51 \\
\bottomrule
\end{tabular}
\end{table}

\subsubsection{Empirical privacy along the ROC}
\label{app:empirical-privacy}
\begin{figure*}[h]
  \centering
  \includegraphics[width=\textwidth]{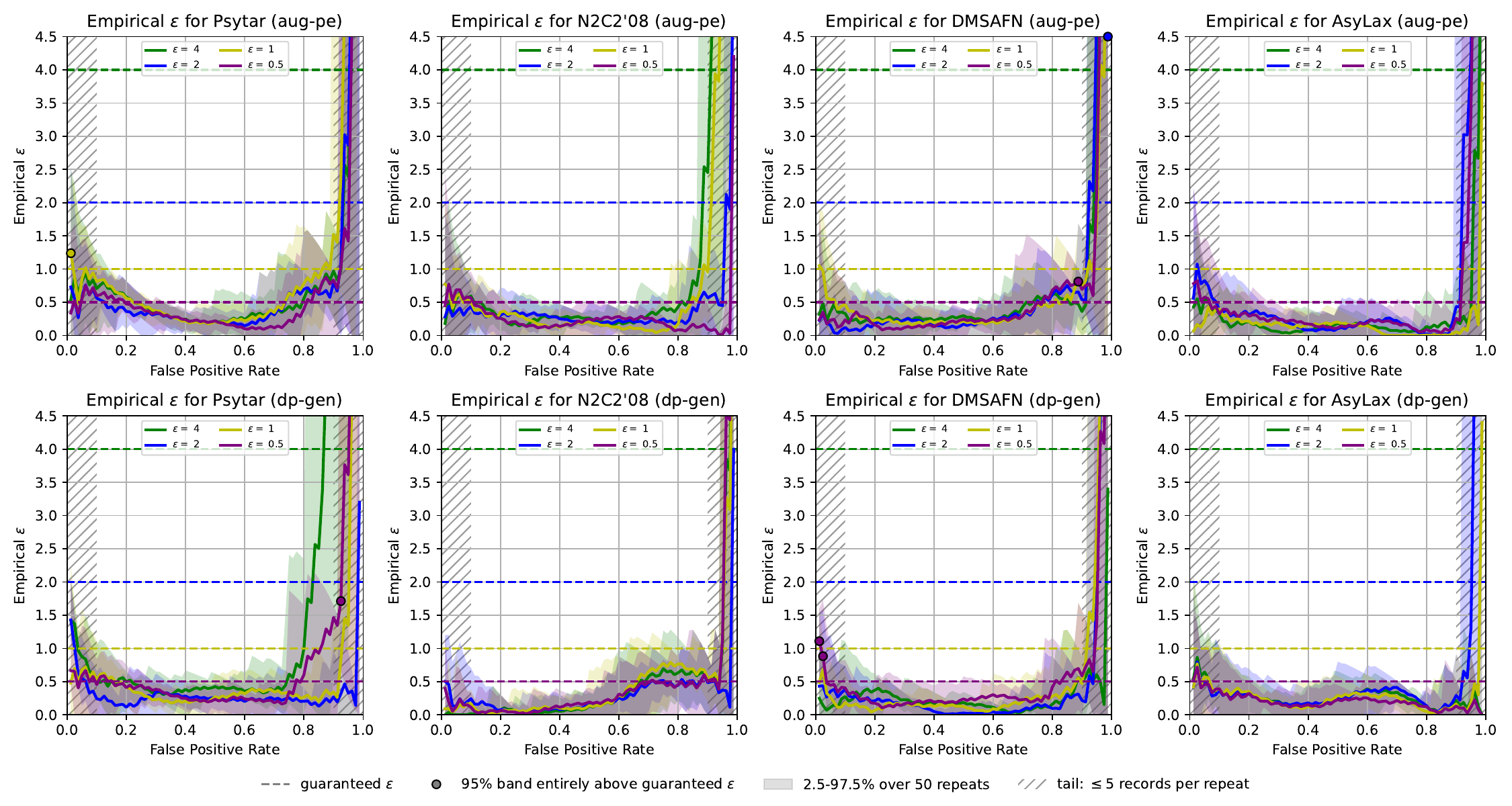}
  \caption{Empirical $\varepsilon$ implied by the bigram reference attack along its ROC (mean over 50 repeats; bands: 2.5--97.5\% quantiles; dashed: guaranteed $\varepsilon$). Hatched: $\alpha<0.1$ or $\alpha>0.9$, where the estimate rests on $\leq$5 records per repeat. Circles: the 2.5\% quantile exceeds the guaranteed $\varepsilon$. }
  \label{fig:eps-emp}
\end{figure*}

We convert the ROC of the bigram reference attack into an empirical privacy level. By the hypothesis-testing characterisation of $(\varepsilon,\delta)$-DP, any membership test with false positive rate $\alpha$ and false negative rate $\beta$ satisfies $\alpha + e^{\varepsilon}\beta \geq 1-\delta$ and $e^{\varepsilon}\alpha + \beta \geq 1-\delta$, so each operating point yields the plug-in estimate
\[
  \hat\varepsilon(\alpha) = \max\Big\{\ln\frac{1-\alpha-\delta}{\beta},\ \ln\frac{1-\beta-\delta}{\alpha},\ 0\Big\},
\]
with $\delta = 1/n^2$ and $n \approx 100$ the number of records scored per repeat.
For each dataset, generator and $\varepsilon$ we run the attack 50 times, each scoring about 50 members and 50 non-members (drawn from a pool of about 150 records), and evaluate $\hat\varepsilon$ on an 81-point grid over $0.01<\alpha<0.99$ (Figure~\ref{fig:eps-emp}). Lines show the mean over repeats, bands the pointwise 2.5--97.5\% quantiles, and dashed lines the guaranteed $\varepsilon$.

Across the central part of the ROC, $\hat\varepsilon$ stays well below the guaranteed $\varepsilon$ for all configurations; in the tails ($\alpha<0.1$ or $\alpha>0.9$, hatched), $\alpha$ or $\beta$ rests on about five records per repeat, and near $\alpha=1$ the estimate diverges as $\beta\to 0$ (we clip it at 10).
We adopt a conservative criterion: an observed exceedance is counted only when the entire band lies above the guaranteed $\varepsilon$, and points where the band reaches the clip are excluded. Under this criterion, we observe exceedances of $\hat\varepsilon$ in 5 of the 32 configurations, including one in the central region of the ROC (DMSAFN, AUG-PE, $\varepsilon=0.5$, at $\alpha=0.89$, by about $0.1$).

\subsubsection{Reference-free Metrics}
\label{app:filtering}
We also measure synthetic text quality using multiple reference-free metrics that reflect linguistic fluency and content diversity. We first compute perplexity and log-likelihood scores for individual samples using a pre-trained GPT-2 model, which indicates how naturally each text reads to the language model. These scores are then aggregated to provide corpus-wide averages. This measures lexical and structural variety through Self-BLEU calculations and Distinct-1/Distinct-2 scores, which capture the percentage of unique single words and word pairs within the corpus. We also generate corpus-level descriptive statistics, including average text length with standard deviation, and assess how well the word frequency distribution matches Zipf's law, a characteristic pattern found in natural language. For texts containing multiple sentences, we evaluate semantic consistency by calculating the average cosine similarity between consecutive sentence embeddings generated by a pre-trained Sentence-BERT model. Additionally, we analyse the emotional tone of the corpus by examining sentiment classification results from a transformer-based model. 

\begin{figure*}[h]
\includegraphics[width=1\textwidth]{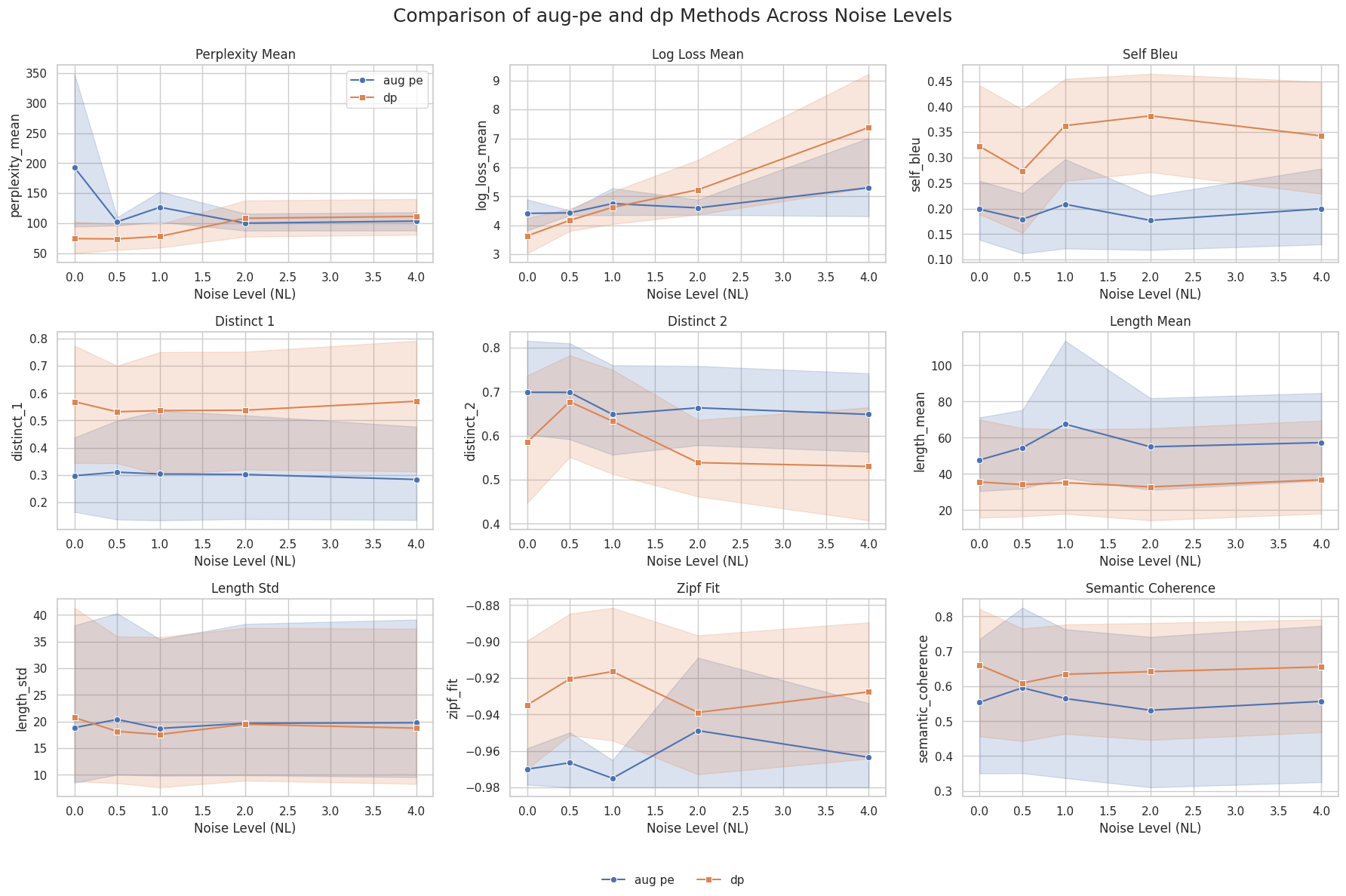}
\caption{Metric trends across epsilon levels for AUG-PE and DP-Gen Methods. Each subplot shows how a specific metric changes with increasing epsilon levels. Solid blue lines represent AUG-PE and dashed orange lines represent DP-Gen, with shaded areas indicating variance across datasets. Metrics are reported across 9 datasets.}
\label{img:ref-free-metrics}
\end{figure*}

Based on the comprehensive evaluation across a variety experiments spanning multiple generation methods, the synthetic text quality shows significant variation depending on the method and dataset combination. As shown in Figure .\ref{img:ref-free-metrics}, the dp-transformers (i.e. DP-Gen) approach consistently produces the most fluent synthetic text with perplexity scores as low as 50.09, while maintaining reasonable diversity, particularly excelling on specialized datasets like \textsc{PsyTAR} and \textsc{DMSAFN}. However, there's a clear trade-off between different quality dimensions: methods that achieve high fluency (low perplexity) often show higher self-similarity, while approaches optimised for diversity may sacrifice semantic coherence. The \textsc{PsyTAR} dataset demonstrates that domain-specific synthetic generation can achieve excellent semantic coherence (\~0.80), whereas more general datasets like Hallmarks struggle with coherence (\~0.08 to 0.12), suggesting that synthetic data quality is heavily influenced by the underlying domain structure. Privacy-preserving noise levels show diminishing returns, with moderate noise (1-2) providing the best balance between privacy protection and text quality, while extreme noise levels (4) significantly degrade fluency without proportional gains in diversity.

\subsection{Additional Implementation Details}
\label{app:additional-implementation}
\noindent\textbf{Computing infrastructure:} All experiments were conducted on a workstation equipped with L40S GPU, Intel Xeon Silver 4216 CPU, and 768~GB RAM, running Ubuntu 22.04 LTS. Additional details are supplied below.

\subsubsection{AUG-PE}
For AUG-PE, we embed documents with ``sentence-t5-base'' and ``kamalkraj/BioSimCSE-BioLinkBERT-BASE" for non-medical datasets and medical datasets, respectively. We use "allenai/OLMoE-1B-7B-0125" for MAUVE calculations, to alleviate the length limit of the model used in the original implementation. To collect named entities, we use "dslim/bert-base-NER" for non-medical datasets and "blaze999/Medical-NER" for medical datasets.

We utilized the TGI (Text Generation Inference)\footnote{https://huggingface.co/docs/text-generation-inference/en/index} framework to accelerate the inference speed of AUG-PE on \textbf{long text datasets}. By leveraging TGI's optimized architecture, we observed a significant reduction in processing time, enabling more efficient experimentation and deployment. 
The Table .\ref{tab:text-tgi-utility} below presents a comparative analysis of utility scores for the \textsc{Psytar} dataset, these results demonstrate that TGI not only enhances inference speed but also maintains the quality of generated outputs.

\begin{table}[ht]
\resizebox{1\columnwidth}{!}{\begin{tabular}{cccc}
\toprule
\midrule
& \textsc{Dataset PsyTAR} (\textbf{Random}: 25.4, \textbf{Majority}: 41.8) & \\
\midrule
\rowcolor{black!10}
\textbf{BERT-base ((original=79.7)} & \textbf{BERT-large (original=80.4)} & \textbf{DeBERTa-xlarge ((original=82.1)} \\
\begin{tabular}{ccc}
Method & $\epsilon = \infty$ & $\epsilon = 2$ \\
AUG-PE-tgi  & 63.0 & 53.8 \\
AUG-PE   & 64.2 & 54.7 \\
\end{tabular}
&
\begin{tabular}{ccc}
Method & $\epsilon = \infty$ & $\epsilon = 2$ \\
AUG-PE-tgi   & 63.7 & 59.6 \\
AUG-PE   & 66.1 & 58.9 \\
\end{tabular}
&
\begin{tabular}{ccc}
Method & $\epsilon = \infty$ & $\epsilon = 2$ \\
AUG-PE-tgi   & 66.9 & 58.7 \\
AUG-PE   & 67.3 & 58.9 \\
\end{tabular} \\
\midrule
\end{tabular}}
\centering
\small
\caption{Comparison of utility scores on the \textsc{Psytar} dataset between the original framework and the TGI framework.
}
\label{tab:text-tgi-utility}
\end{table}

\paragraph{Bug Report}
During our experiments with AUG-PE, we identified and reported a bug related to the use of the "sentence-t5-base" model as a feature extractor for calculating embedding distances in the voting histogram. Specifically, we observed that when the index was moved from CPU to GPU, the resulting calculations were incorrect, yielding all-zero distances for most samples. However, when the index remained on the CPU, the calculations proceeded as expected and produced reliable results. To ensure the validity and reliability of our outcomes when using the "sentence-t5-base" model, we commented out the two lines of code responsible for transferring the index to the GPU (in ``./dpsda/dp\_counter.py", line 30-31). This workaround allows us to maintain accurate distance computations until the underlying issue is resolved.

\subsubsection{DP-Gen}
Owing to memory limitations, we employed flash-attention\footnote{https://github.com/Dao-AILab/flash-attention} to facilitate experimentation on datasets with extended context lengths. To ensure compatibility with flash-attention, we modified the Opacus codebase (specifically the optimizer), casting the noise added to the gradients to match the data type of the gradients.

\subsubsection{EPSVec}
We run EPSVec from the authors' released code with the settings below, which are
the same for all datasets unless stated otherwise.

\begin{itemize}
  \item \textbf{Model.} Llama-3.2-1B (base) is used to extract the steering
  vectors, to generate the synthetic data, and to produce the public pool of
  few-shot candidates.

  \item \textbf{Classes.} Each label combination is a class. Combinations seen
  fewer than twice are merged into a single class \emph{other}, whose generated
  records take the label sets of the real records they replace. Records with no
  label form the class \emph{none}. For \textsc{N2C2'08}, classes are defined on
  its three most frequent labels. \footnote{EPSVec conditions generation on a class, which for a multi-label dataset is a
combination of labels. N2C2'08 has 16 labels whose combinations are mostly too rare to
condition on, so we define classes by the three most frequent labels (Hypertension,
Diabetes, CAD): every training record falls into exactly one of the $2^3 = 8$ combinations
of these labels, including the one with none of them. All records are thus covered, and
we generate one note per record from its combination. To evaluate on the same 16-label
task as the other methods, we then shuffle the full label sets of the real records in
each combination and assign them at random to the notes generated from that combination.
Each note thus keeps the three labels it was generated from, while its other 13 labels
carry only their distribution given those three, not information about the note's text.}

  We generate exactly as many records per class as the real training set
  contains, so every synthetic set matches its real counterpart in size and
  label distribution.

  \item \textbf{Privacy.} The total budget $\varepsilon$ is split into
  $\varepsilon/6$ ($\delta = 10^{-6}$) for privately selecting two few-shot
  examples per class and $5\varepsilon/6$ ($\delta = 9\times10^{-6}$) for
  privatising the steering vectors, i.e.\ $\delta = 10^{-5}$ in total. The noised
  vectors are drawn once per (dataset, $\varepsilon$) and reused for every
  generated record, so generation spends no further privacy budget.

  \item \textbf{Few-shot candidates.} $\min(\max(n_c/2, 10), 50)$ candidates per
  class, where $n_c$ is the number of training records in the class, kept if
  Qwen3-4B-Instruct-2507 rates them at least 6 on a 0--10 quality scale. The number of few-shot candidates was established empirically by minimising the probability of the noise added to the histogram comparison to private data (which scales with the number of candidates) resulting in randomly drawing a candidate that has received zero votes from private data. 
  Embeddings are computed with Qwen3-Embedding-8B.

  \item \textbf{Generation.} The steering vector is added at layers 8--11 (the
  default location in the released code), with temperature 1.6, top-$p$ 0.9,
  top-$k$ 50 and seed 42. Outputs that the same judge scores below 2 are
  discarded and regenerated. The maximum number of new tokens is 80 for
  \textsc{ZsTFNS}, 128 for \textsc{PsyTar} and \textsc{HoC}, 160 for
  \textsc{DMSAFN}, 2048 for \textsc{Lc37Fnd} and \textsc{EurLex}, 3072 for
  \textsc{N2C2'08}, 4500 for \textsc{MIMIC}, and 3610 for \textsc{AsyLax}
  (the 95th percentile of its real texts' length).

  \item \textbf{Injection coefficient.} We tune the coefficient that scales the
  steering vector separately for each (dataset, $\varepsilon$) over
  $\{0.2, 0.4, 0.7, 1.0, 1.4\}$ (1.4 is the released default), on 300 generated
  samples per value. We choose the value such that the
  share of samples that a ModernBERT-base classifier trained on the real training
  set assigns to their intended class. The
  exception is \textsc{AsyLax}: at 1.4 its long documents degenerate into
  whitespace (91--99\% of characters, against 21\% in real texts), and only a low
  coefficient prevented this, so we use 0.2 for every $\varepsilon$. We note that such hyper-parameter optimisation is not covered by the privacy accounting.
\end{itemize}

\subsection{DP-Postprocessing: Filtering}
\label{app:post-processing}
The data filtering post-processing method employed in this study is both straightforward and effective. A domain-specific prompt is presented to a large language model (LLM); for example, in the case of the Psytar dataset, the prompt may inquire whether a given text resembles a social media post discussing medication side effects. The LLM is then tasked with generating a single token in response. The final filtering score for each sample is computed as the difference between the model-assigned probabilities of the "yes" and "no" tokens, denoted as $P(yes)$-$P(no)$. This scoring mechanism provides a quantitative measure of the model’s confidence in classifying the text according to the specified prompt, thereby enabling efficient filtering of relevant data samples. The following Table .\ref{tab:filtering_results} presents the improvements on Mauve score after applying filtering method. These results indicate that the quality of synthetic datasets can be further improved by applying post-processing filtering techniques, suggesting a promising direction for future research.
\begin{table}[ht]
\centering
\begin{tabular}{lcccc}
\hline
\textbf{Dataset} & \textbf{$\epsilon=4$} & \textbf{$\epsilon=2$} & \textbf{$\epsilon=1$} & \textbf{$\epsilon=0.5$} \\
\hline
\textsc{Psytar} & -0.0598 & -0.0317 & -0.0119 & 0.0055  \\
\textsc{DMSAFN} & 0.034 & 0.0406 & 0.0096 & 0.0191  \\
\hline
\end{tabular}
\caption{Improvements in Mauve scores for different $\epsilon$ values across datasets after the application of post-processing filtering using DP-Gen.}
\label{tab:filtering_results}
\end{table} 


\end{document}